\documentclass[10pt,journal,compsoc]{IEEEtran}
\usepackage{amsmath}
\usepackage{amssymb}
\usepackage{booktabs}
\usepackage{epsfig}
\usepackage{enumitem}
\usepackage{multirow}
\usepackage{setspace}
\usepackage{microtype}      
\usepackage[american]{babel}
\usepackage{url}
\usepackage{bm}
\usepackage{color}
\newcommand{\eg}{\emph{e.g.}}

\newcommand{\ie}{\emph{i.e.}}
\newcommand{\etal}{\emph{et al.}}

\graphicspath{{figure/}}

\usepackage{color}

\ifCLASSOPTIONcompsoc
  \usepackage[nocompress]{cite}
\else
  \usepackage{cite}
\fi

\ifCLASSINFOpdf
\else
\fi

\begin{document}
%

\title{Fine-Grained Image Analysis with Deep Learning: A Survey}

\author{
	Xiu-Shen Wei,~\IEEEmembership{Member,~IEEE}, Yi-Zhe Song,~\IEEEmembership{Senior Member,~IEEE}, Oisin Mac Aodha, Jianxin Wu,~\IEEEmembership{Member,~IEEE}, Yuxin Peng,~\IEEEmembership{Senior Member,~IEEE}, Jinhui Tang,~\IEEEmembership{Senior Member,~IEEE}, Jian Yang,~\IEEEmembership{Member,~IEEE}, Serge Belongie
\IEEEcompsocitemizethanks{\IEEEcompsocthanksitem {\scriptsize X.-S. Wei and J. Yang are with PCA Lab, Key Lab of Intelligent Perception and Systems for High-Dimensional Information of Ministry of Education, and Jiangsu Key Lab of Image and Video Understanding for Social Security, School of Computer Science and Engineering, Nanjing University of Science and Technology, China. Y.-Z. Song is with University of Surrey, UK. O. Mac Aodha is with the University of Edinburgh, UK. J. Wu is the State Key Laboratory for Novel Software Technology, Nanjing University, China. Y. Peng is with Peking University, China. J. Tang is with Nanjing University of Science and Technology, China. S. Belongie is with the University of Copenhagen and the Pioneer Centre for AI, Denmark.}
\IEEEcompsocthanksitem {\scriptsize X.-S. Wei and J. Yang are the corresponding authors.}}
}

\markboth{ACCEPTED BY IEEE TPAMI}%
{Wei \MakeLowercase{\textit{et al.}}: Fine-Grained Image Analysis with Deep Learning: A Survey}
%

\IEEEtitleabstractindextext{%
\begin{abstract}
Fine-grained image analysis (FGIA) is a longstanding and fundamental problem in computer vision and pattern recognition, and underpins a diverse set of real-world applications. The task of FGIA targets analyzing visual objects from subordinate categories, \eg, species of birds or models of cars. The small inter-class and large intra-class variation inherent to fine-grained image analysis makes it a challenging problem. Capitalizing on advances in deep learning, in recent years we have witnessed remarkable progress in deep learning powered FGIA. In this paper we present a systematic survey of these advances, where we attempt to re-define and broaden the field of FGIA by consolidating two fundamental fine-grained research areas -- fine-grained image recognition and fine-grained image retrieval. In addition, we also review other key issues of FGIA, such as publicly available benchmark datasets and related domain-specific applications. We conclude by highlighting several research directions and open problems which need further exploration from the community.
\end{abstract}

}

\maketitle


\IEEEdisplaynontitleabstractindextext

%
\IEEEpeerreviewmaketitle

\section{Introduction}



\begin{figure}[t!]
\centering
{\includegraphics[width=\columnwidth]{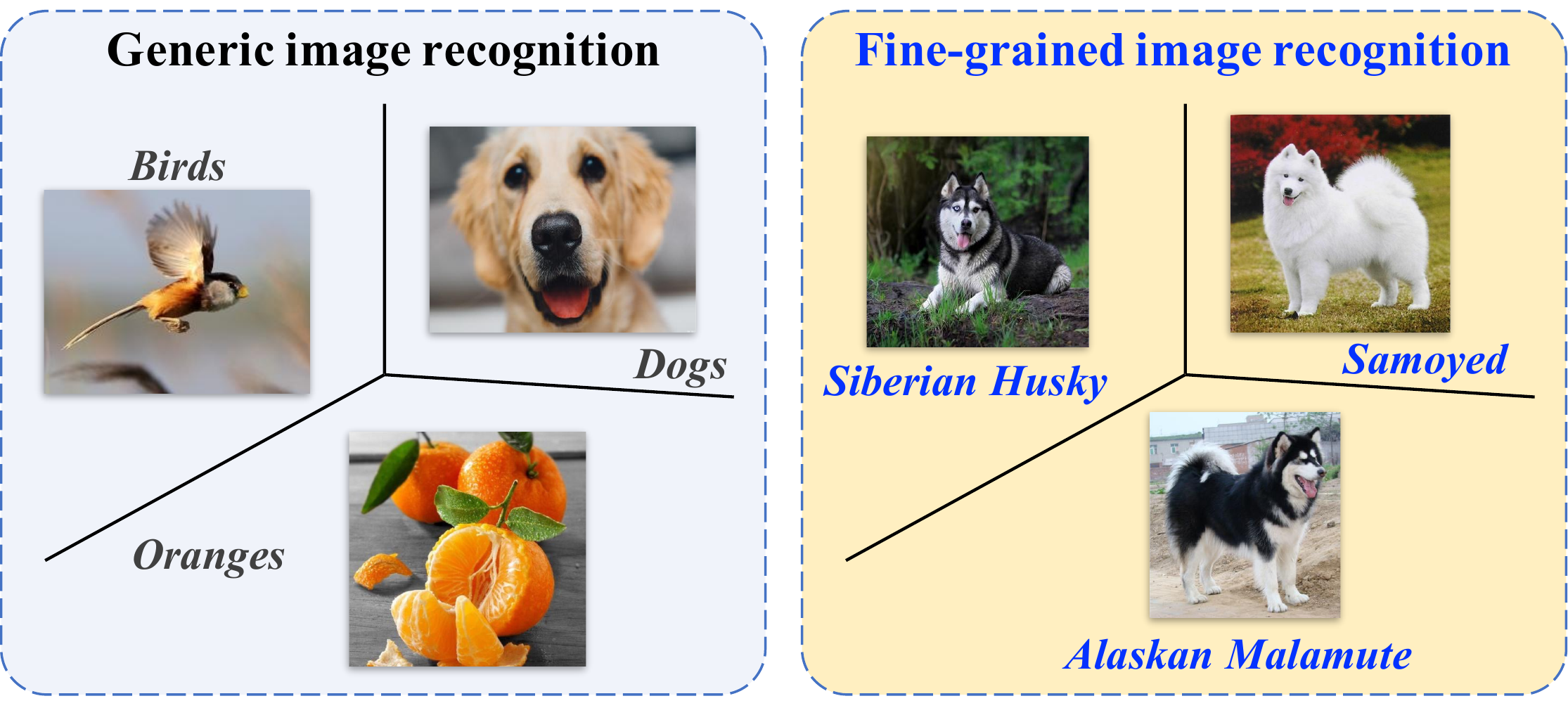}}
\vspace{-20pt}
\caption{Fine-grained image analysis \emph{vs}. generic image analysis (using visual classification as an example).
}
\label{fig:fgvsgeneric}
\end{figure}

\IEEEPARstart{T}{he} human visual system is inherently capable of fine-grained image reasoning -- we are not only able to tell a dog from a bird, but also know the difference between a Siberian Husky and an Alaskan Malamute (see Figure~\ref{fig:fgvsgeneric}). Fine-grained image analysis (FGIA) was introduced to the academic community for the very same purpose, \ie, to teach machine to ``see'' in a fine-grained manner. FGIA approaches are present in a wide-range of applications in both industry and research, with examples including automatic biodiversity monitoring~\cite{nabirds15,inat2017,inat2021}, intelligent retail~\cite{singleExFG2017cvpr,rpc,jia2020fashionpedia}, and intelligent transportation~\cite{cviucarreid,FGREIDIJCV2014}, and have resulted in a positive impact in areas such as conservation~\cite{FGICCVworkshop} and commerce~\cite{retailDL}.

The goal of FGIA in computer vision is to retrieve and recognize images belonging to multiple subordinate categories of a super-category (\emph{aka} a meta-category or a basic-level category), \eg, different species of animals/plants, different models of cars, different kinds of retail products, etc. 
The key challenge therefore lies with understanding fine-grained visual differences that sufficiently discriminate between objects that are highly similar in overall appearance, but differ in \emph{fine-grained} features. Great strides has been made since its inception almost two decades ago~\cite{Johnsoncognitive,feifeiFG2010,WahCUB200_2011}.
Deep learning~\cite{natureDL} in particular has emerged as a powerful method for discriminative feature learning, and has led to remarkable breakthroughs in the field of FGIA. Deep learning enabled FGIA has greatly advanced the practical deployment of these methods in a diverse set of application scenarios~\cite{cviucarreid,FGREIDIJCV2014,FGICCVworkshop,rpc}.


\begin{figure*}[t!]
\centering
{\includegraphics[width=0.85\textwidth]{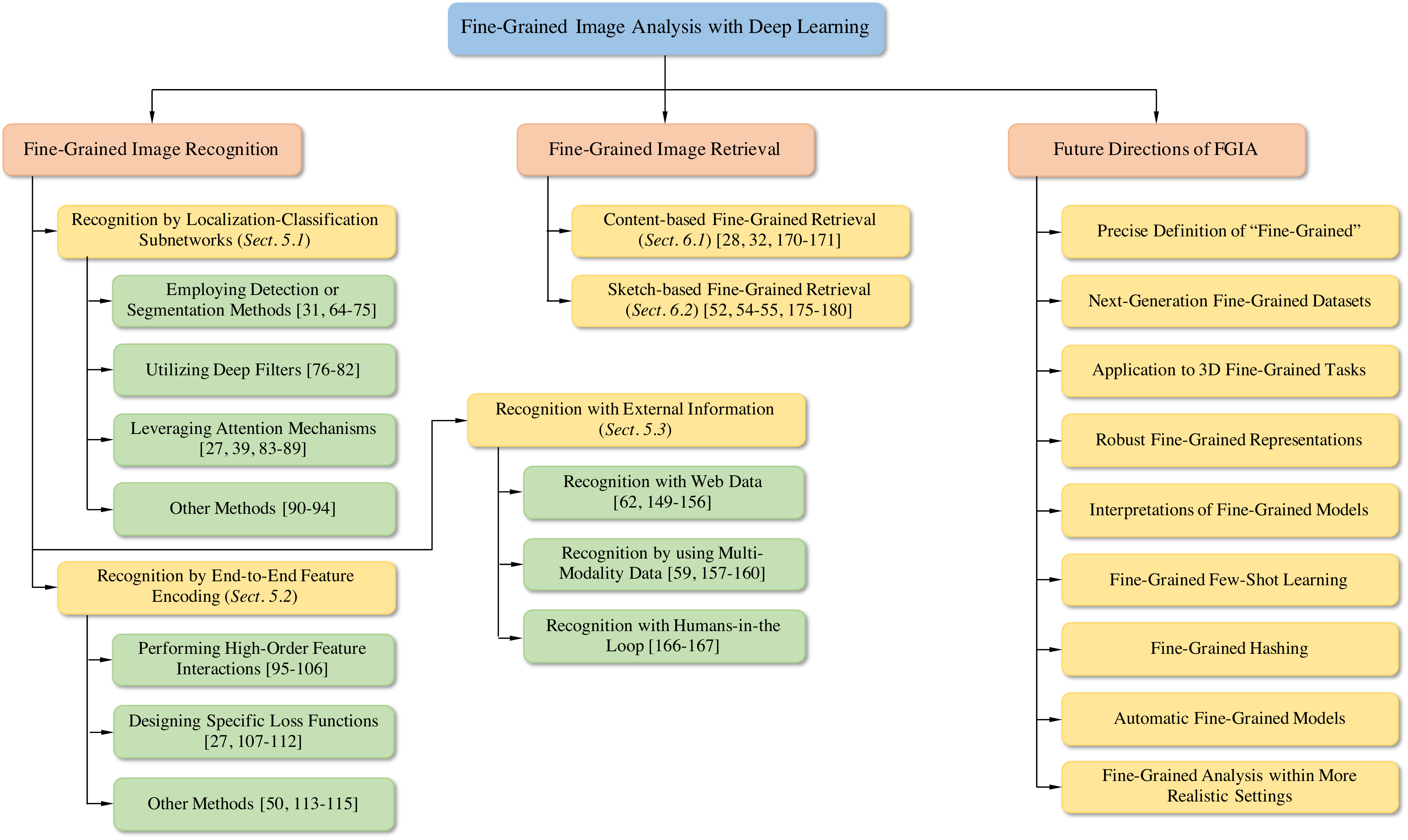}}
\vspace{-10pt}
\caption{Overview of the landscape of deep learning based fine-grained image analysis (FGIA), as well as future directions.
}
\label{fig:structure}
\end{figure*}

There has been significant interest in FGIA from both the computer vision and machine learning research communities in recent years. Rough statistics indicate an average of $>10$ conference papers on deep learning based FGIA are published every year in each of the premium vision and machine learning conferences. 
There have also been a number of special issues addressing FGIA~\cite{tpamiSI,PRSIFG,TOMMSI,PRLSI,NUMSI}.
Additionally, a number of influential competitions on FGIA are frequently held on online platforms. 
Representatives include the series of iNaturalist Competitions (for large numbers of natural species)~\cite{inatcomp}, the Nature Conservancy Fisheries Monitoring (for fish species categorization)~\cite{fishcomp}, Humpback Whale Identification (for whale identity categorization)~\cite{whalecomp}, among others. Each competition attracted hundreds of participants from around the world, and some even exceeding 2,000 teams. Specific tutorials and workshops aimed at FGIA topics have also been held at top-tier international conferences, \eg,~\cite{CVPRtutorial,CVPRworkshop}.

Despite such salient interest, the study of FGIA with deep learning remains fragmented. It is therefore the purpose of this survey to (i) provide a comprehensive survey of recent achievements in FGIA, especially those brought by deep learning techniques, and more importantly (ii) to propose a unified research front by consolidating research from different aspects of FGIA. Our approach is significantly different to existing surveys~\cite{bozhao_ijac_2017,survey18ICSP} that focus solely on the problem of fine-grained \emph{recognition/classification}, which as we argue only constitutes part of the larger study of FGIA. In particular, we attempt to re-define and broaden the field of fine-grained image analysis, by highlighting the synergy between fine-grained \textit{recognition}, and the parallel but complementary task of fine-grained image \emph{retrieval}, which is also an integral part of FGIA.



Our survey takes a unique deep learning based perspective to review recent advances in FGIA in a broad, systematic, and comprehensive manner. Our main contributions are summarized as follows:
\begin{itemize}[leftmargin = 2.2em]
\item We broaden the field of FGIA, by offering a consolidated landscape that promotes synergies between related problems in fine-grained image analysis. 
\item We provide a comprehensive review of FGIA techniques based on deep learning, including commonly accepted problem definitions, benchmark datasets, different families of FGIA methods, along with covering domain-specific FGIA applications.
Particularly, we organize these approaches taxonomically (see Figure~\ref{fig:structure}) to provide readers with a quick snapshot of the state-of-the-art in this area.
\item We consolidate the performance of existing methods on several publicly available datasets 
and provide discussion and insights to inform future research.
\item We finish by discussing existing challenges and open issues, and identify new trends and future directions to provide a plausible road map for the community to address these problems.
\item Finally, in an attempt to continuously track recent developments in this fast advancing field, we provide an accompanying webpage which catalogs papers addressing FGIA problems, according to our problem-based taxonomy: \url{http://www.weixiushen.com/project/Awesome_FGIA/Awesome_FGIA.html}.
\end{itemize}


\section{Recognition vs. Retrieval}\label{sec:tax}
Previous surveys of FGIA, \eg,~\cite{bozhao_ijac_2017,survey18ICSP}, predominately focused on fine-grained recognition, and as a result do not expose all facets of the FGIA problem. 
In this survey, we cover two fundamental areas of fine-grained image analysis for the first time (\ie, recognition and retrieval) in order to give a comprehensive review of recent advances in deep learning based FGIA techniques. 
In Figure~\ref{fig:structure}, we provide a new taxonomy that reflects the current FGIA landscape.

\textbf{Fine-Grained Recognition:} 
We organize the different families of fine-grained \emph{recognition} approaches into three paradigms, \ie, 1) recognition by localization-classification subnetworks, 2) recognition by end-to-end feature encoding, and 3) recognition with external information. 
Fine-grained recognition is the most studied area in FGIA, since recognition is a fundamental ability of most visual systems and is thus worthy of long-term continuous research.

\textbf{Fine-Grained Retrieval:} 
Based on the type of query image, we separate fine-grained \emph{retrieval} methods into two groups, \ie, 1) content-based fine-grained image retrieval and 2) sketch-based fine-grained image retrieval. Compared with fine-grained recognition, fine-grained retrieval is an emerging area of FGIA in recent years, one that is attracting more and more attention from both academia and industry.

\textbf{Recognition and Retrieval Differences:} Both fine-grained recognition and retrieval aim to identify the discriminative, but subtle, differences between different fine-grained objects. 
However, fine-grained recognition is a closed-world task with a \emph{fixed} number of subordinate categories. 
In contrast, fine-grained retrieval extends the problem to an open-world setting with unlimited sub-categories. Furthermore, fine-grained retrieval also aims to rank all the instances so that images depicting the concept of interest (\eg, the same sub-category label) are ranked highest based on the fine-grained details in the query.

\textbf{Recognition and Retrieval Synergies:} Advances in fine-grained recognition and retrieval have commonalities and can benefit each other. Many common techniques are shared by both fine-grained recognition and retrieval, \eg, deep metric learning methods~\cite{MAMCeccv,xiawuaaai19}, multi-modal matching methods~\cite{WACV2020fgictextvision,wacv2021visiontest}, and the basic ideas of selecting useful deep descriptors~\cite{maskcnnPR,Wei16scda}, etc. Detailed discussions are elaborated in Section~\ref{sec:commontech}. Furthermore, in real-world applications, fine-grained recognition and retrieval also compliment each other, \eg, retrieval techniques are able to support novel sub-category recognition by utilizing learned representations from a fine-grained recognition model~\cite{DeepFashion2,rpc}.

\section{Background: Problem and Challenges}\label{sec:background}

\begin{figure}[t!]
\centering
{\includegraphics[width=0.95\columnwidth]{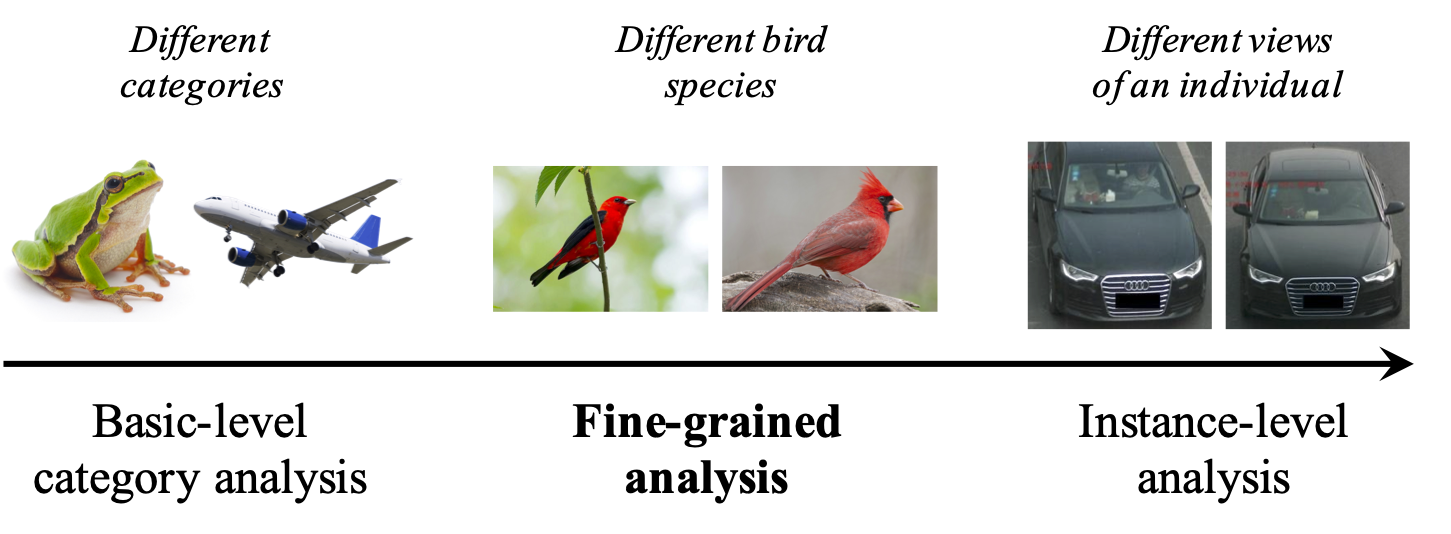}}
\vspace{-1em}
\caption{An illustration of fine-grained image analysis which lies in the continuum between the basic-level category analysis (\ie, generic image analysis) and the instance-level analysis (\eg, car identification).}
\label{fig:multigrained}
\end{figure}

Fine-grained image analysis (FGIA) focuses on dealing with objects belonging to multiple \emph{subordinate categories} of the same meta-category (\eg, different species of birds or different models of cars), and generally involves two central tasks: fine-grained image recognition and fine-grained image retrieval. As illustrated in Figure~\ref{fig:multigrained}, fine-grained analysis lies in the continuum between basic-level category analysis (\ie, generic image analysis) and instance-level analysis (\eg, the identification of individuals).

\begin{figure}[b]
\centering
{\includegraphics[width=0.95\columnwidth]{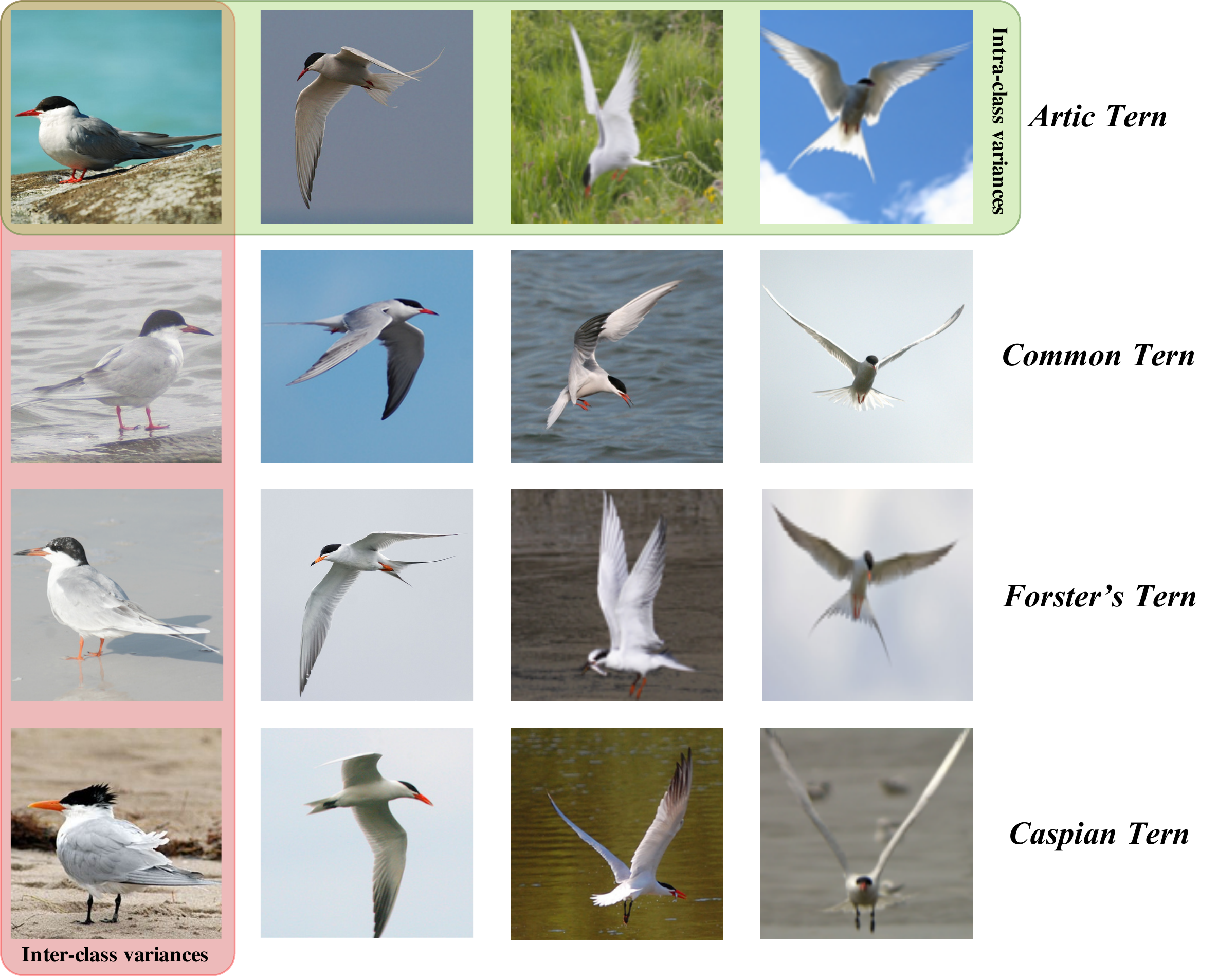}}
\vspace{-1em}
\caption{Key challenges of fine-grained image analysis, \ie, small inter-class variations and large intra-class variations. Here we present four different \texttt{Tern} species from \cite{WahCUB200_2011}, one species per row, with different instances in the columns.}
\label{fig:cub_example}
\end{figure}

Specifically, what distinguishes FGIA from generic image analysis is that in generic image analysis, target objects belong to coarse-grained meta-categories (\ie, basic-level categories) and are thus visually quite different (\eg, determining if an image contains a bird, a fruit, or a dog). However, in FGIA, since objects typically come from sub-categories of the same meta-category, the fine-grained nature of the problem causes them to be visually similar. As an example of fine-grained recognition, in Figure~\ref{fig:fgvsgeneric}, the task is to classify different breeds of dogs. For accurate image recognition, it is necessary to capture the subtle visual differences (\eg, discriminative features such as ears, noses, or tails). Characterizing such features is also desirable for other FGIA tasks (\eg, retrieval). Furthermore, as noted earlier, the fine-grained nature of the problem is challenging because of the \emph{small inter-class variations} caused by highly similar sub-categories, and the \emph{large intra-class variations} in poses, scales and rotations (see Figure~\ref{fig:cub_example}). It is as such the opposite of generic image analysis (\ie, the small intra-class variations and the large inter-class variations), and what makes FGIA a unique and challenging problem.


While instance-level analysis typically targets a \emph{specific instance} of an object not just object categories or even object sub-categories, if we move down the spectrum of granularity, in the extreme, individual identification (\eg, face identification) can be viewed as a special instance of fine-grained recognition, where the granularity is at the \textit{individual} identity level. For instance, person/vehicle re-identification~\cite{reidsurveyarxiv,cviucarreid} can be considered a fine-grained task, which aims to determine whether two images are taken of the same specific person/vehicle. In practice, these works solve the corresponding domain-specific problems using related methods to FGIA, \eg, by capturing the discriminative parts of objects (faces, people, and vehicles)~\cite{facenet,FGREIDIJCV2014,personreideccv18fg}, discovering coarse-to-fine structural information~\cite{rnnha}, developing attribute-based models~\cite{deepfashion16,yuanqingaaai17local}, and so on. Research in these instance-level problems is also very active. However, since such problems are not within the scope of classical FGIA (see Figure~\ref{fig:multigrained}), for more information, we refer readers to survey papers of these specific topics, \eg,~\cite{reidsurveyarxiv,cviucarreid,deepfacesurvey}. In the following, we start by formulating our definition of classical FGIA.

\textbf{Formulation:} In generic image recognition, we are given a training dataset $\mathcal{D}=\left\{ \left( \bm{x}^{(n)}, y^{(n)}\right) | i=1, ..., N  \right\}$, containing multiple images and associated class labels (\ie, $\bm{x}$ and $y$), where $y\in[1, ..., C]$. 
Each instance $\left(\bm{x}, y\right)$ belongs to the joint space of both the image and label spaces (\ie, $\mathcal{X}$ and $\mathcal{Y}$, respectively), according to the distribution of $p_r(\bm{x}, y)$
\begin{equation}
\left(\bm{x}, y\right) \in \mathcal{X}\times \mathcal{Y}\,.
\end{equation}
In particular, the label space $\mathcal{Y}$ is the union space of all the $C$ subspaces corresponding to the $C$ categories, \ie, $\mathcal{Y} = \mathcal{Y}_1 \cup \mathcal{Y}_2 \cup \cdots \cup \mathcal{Y}_c \cup \cdots \cup \mathcal{Y}_C$. Then, we can train a predictive/recognition deep network $f(\bm{x};\theta)$ parameterized by $\theta$ for generic image recognition by minimizing the expected risk
\begin{equation}
\min_\theta \mathbb{E}_{(\bm{x},y)\sim p_r(\bm{x}, y)} \left[ \mathcal{L}(y, f(\bm{x};\theta))\right]\,,
\end{equation}
where $\mathcal{L}(\cdot,\cdot)$ is a loss function that measures the match between the true labels and those predicted by $f(\cdot;\theta)$. 
While, as aforementioned, fine-grained recognition aims to accurately classify instances of different subordinate categories from a certain meta-category, \ie,
\begin{equation}
\left(\bm{x}, y'\right) \in \mathcal{X}\times \mathcal{Y}_c\,,
\end{equation}
where $y'$ denotes the fine-grained label and $\mathcal{Y}_c$ represents the label space of class $c$ as the meta-category. Therefore, the optimization objective of fine-grained recognition is as
\begin{equation}
\min_\theta \mathbb{E}_{(\bm{x},y')\sim p'_r(\bm{x}, y')} \left[ \mathcal{L}(y', f(\bm{x};\theta))\right]\,.
\end{equation}

Compared with fine-grained recognition, in addition to getting the sub-category correct, fine-grained retrieval also must rank all the instances so that images belonging to the same sub-category are ranked highest based on the fine-grained details in the query of retrieval tasks. 
Given an input query $\bm{x}^q$, the goal of a fine-grained retrieval system is to rank all instances in a retrieval set $\Omega=\{\bm{x}^{(i)}\}_{i=1}^M$ (whose label $y'\in \mathcal{Y}_c$) based on their fine-grained relevance to the query. Let $\mathcal{S}_\Omega = \{s^{(i)}\}_{i=1}^M$ represent the similarity between $\bm{x}^q$ and each  $\bm{x}^{(i)}$ measured via a pre-defined metric applied to the corresponding fine-grained representations, \ie, $h(\bm{x}^q; \delta)$ and $h(\bm{x}^{(i)}; \delta)$. 
Here, $\delta$ denotes the parameters of a retrieval model $h$. 
For the instances whose labels are consistent with the fine-grained category of $\bm{x}^q$, we form them into a positive set $\mathcal{P}_q$ and obtain the corresponding $\mathcal{S}_P$. Then, the retrieval model $h(\cdot; \delta)$ can be trained by maximizing the ranking based score 
\begin{equation}
\max_\delta \frac{\mathcal{R}(i, \mathcal{S}_P)}{\mathcal{R}(i, \mathcal{S}_\Omega)}\,,
\end{equation}
w.r.t. all the query images, where $\mathcal{R}(i,\mathcal{S}_P)$ and $\mathcal{R}(i, \mathcal{S}_\Omega)$ refer to the rankings of the instance $i$ in $\mathcal{P}_q$ and $\Omega$, respectively.


\section{Benchmark Datasets}\label{sec:datasets}

In recent years, the vision community has released many fine-grained benchmark datasets covering diverse domains, \eg, birds~\cite{WahCUB200_2011,Birdsnap14,nabirds15}, dogs~\cite{Khosla11stanforddogs,MAMCeccv}, cars~\cite{cars}, airplanes~\cite{airplanes}, flowers~\cite{Flowers08}, vegetables~\cite{vegfru}, fruits~\cite{vegfru}, foods~\cite{food101}, fashion~\cite{deepfashion16,DeepFashion2,jia2020fashionpedia}, retail products~\cite{rpc,product10k}, etc. Additionally, it is worth noting that even the most popular large-scale image classification dataset, \ie, ImageNet~\cite{imgnet}, also contains fine-grained classes covering a lot of dog and bird sub-categories.

\begin{figure*}[t]
\centering
{\includegraphics[width=0.95\textwidth]{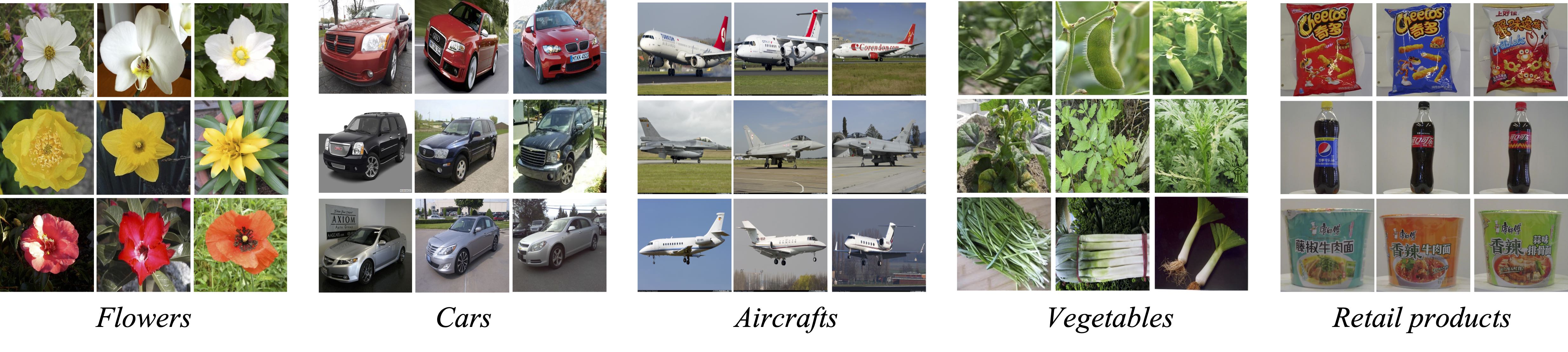}}
\vspace{-1.8em}
\caption{Examples of fine-grained images belonging to different species of flowers/vegetables~\cite{vegfru}, different models of cars~\cite{cars} and aircraft~\cite{airplanes} and different kinds of retail products~\cite{rpc}. Accurate identification of these fine-grained objects requires the extraction of discriminative, but subtle, object parts or image regions. (Best viewed in color and zoomed in.)}
\label{fig:datasetdemo}
\end{figure*}

\begin{table*}[t]
\centering
\scriptsize
\caption{Summary of popular fine-grained image datasets organized by their major applicable topics and sorted by their release time. Note that, ``$\sharp$ images'' means the total number of images of these datasets. ``BBox'' indicates whether this dataset provides object bounding box supervisions. ``Part anno.'' means that key parts annotations are provided. ``HRCHY'' corresponds to hierarchical labels. ``ATR'' represents attribute labels (\eg, wing color, male, female, etc). ``Texts'' indicates whether fine-grained text descriptions of images are supplied. Several datasets are listed here twice since they are commonly used in both recognition and retrieval tasks.}
\vspace{-1em}
	\begin{tabular}{|c|r||c|c|r|c|c|c|c|c|c|}
		\hline
		Topic & Dataset name~~~~~ & Year & Meta-class & $\sharp$ images~~~~ & $\sharp$ categories  & BBox & Part anno. & HRCHY& ATR & Texts \\
		\hline \hline 
		\multirow{15}{*}{Recog.} & \textit{Oxford Flowers}~\cite{Flowers08} & 2008	 & Flowers & ~~~~8,189 & ~~~~102  &  &  & &  & \checkmark \\ \cline{2-11}
		& \textit{CUB200-2011}~\cite{WahCUB200_2011} & 2011 & Birds   & ~~11,788 & ~~~~200   &\checkmark & \checkmark &  & \checkmark & \checkmark\\ \cline{2-11}
		& \textit{Stanford Dogs}~\cite{Khosla11stanforddogs} & 2011& Dogs & ~~20,580 & ~~~~120   &\checkmark & &  &   & \\ \cline{2-11}
		& \textit{Stanford Cars} ~\cite{cars} & 2013 & Cars & ~~16,185 & ~~~~196   &\checkmark &  & &  & \\ \cline{2-11}
		& \textit{FGVC Aircraft}~\cite{airplanes}	& 2013 & Aircrafts  & ~~10,000 & ~~~~100    &\checkmark & &\checkmark &  & \\ \cline{2-11}
		& \textit{Birdsnap}~\cite{Birdsnap14} & 2014 & Birds & ~~49,829 & ~~~~500   &\checkmark & \checkmark & & \checkmark &  \\ \cline{2-11}
		& \textit{Food101}~\cite{food101} & 2014 & Food dishes & 101,000  &  ~~~~101 & &  & &  &  \\ \cline{2-11}
		& \textit{NABirds}~\cite{nabirds15} & 2015 & Birds & ~~48,562 & ~~~~555   &\checkmark & \checkmark & & &  \\ \cline{2-11}
		& \textit{Food-975}~\cite{BGLCVPR2016} & 2016 & Foods & ~~37,885 & ~~~~975   & &  & & \checkmark &  \\ \cline{2-11}
		& \textit{DeepFashion}~\cite{deepfashion16} & 2016 & Clothes & 800,000 & ~~1,050   &\checkmark & \checkmark & & \checkmark &  \\ \cline{2-11}
		& \textit{Fru92}~\cite{vegfru} & 2017 & Fruits & ~~69,614 & ~~~~~~92   & &  & \checkmark&  &  \\ \cline{2-11}
		& \textit{Veg200}~\cite{vegfru} & 2017 & Vegetable & ~~91,117 & ~~~~200   &&  & \checkmark &  &  \\ \cline{2-11}
		& \textit{iNat2017}~\cite{inat2017} & 2017 & Plants \& Animals & 857,877 & ~~5,089   &\checkmark &  & \checkmark &  &  \\ \cline{2-11}
	    & \textit{Dogs-in-the-Wild}~\cite{MAMCeccv} & 2018 & Dogs & 299,458 & ~~~~362  & &  & &  &  \\ \cline{2-11}
		& \textit{RPC}~\cite{rpc} & 2019 & Retail products & ~~83,739 & ~~~~200  &\checkmark &  & \checkmark &  &  \\ \cline{2-11}
		& \textit{Products-10K}~\cite{product10k} & {2020} & {Retail products} & ~{150,000} & {10,000}  &{\checkmark} &  & {\checkmark} &  &  \\ \cline{2-11}
		& \textit{iNat2021}~\cite{inat2021} & {2021} & {Plants \& Animals} & {3,286,843} & {10,000}  &  &  & {\checkmark} &  &  \\  \hline\hline
		\multirow{11}{*}{Retriev.} & \textit{Oxford Flowers}~\cite{Flowers08} & 2008	 & Flowers & ~~~~8,189 & ~~~~102   &  &  & &  & \checkmark \\ \cline{2-11}
		& \textit{CUB200-2011}~\cite{WahCUB200_2011} & 2011 & Birds   & ~~11,788 & ~~~~200  &\checkmark & \checkmark &  & \checkmark & \checkmark\\ \cline{2-11}
		& \textit{Stanford Cars} ~\cite{cars} & 2013 & Cars & ~~16,185 & ~~~~196   &\checkmark &  & &  & \\ \cline{2-11}
		& \textit{SBIR2014}$^*$~\cite{SBIRbmvc14} & 2014 & Multiple & ~~~1,120/7,267 & ~~~~~14   &\checkmark & \checkmark & & \checkmark &  \\ \cline{2-11}
		& \textit{DeepFashion}~\cite{deepfashion16} & 2016 & Clothes & 800,000 & ~~1,050   &\checkmark & \checkmark & & \checkmark &  \\ \cline{2-11}
		& \textit{QMUL-Shoe}$^*$~\cite{thatshoecvpr} & 2016 & Shoes & 419/419 & ~~~~~~~~1   & &  & & \checkmark &  \\ \cline{2-11}
		& \textit{QMUL-Chair}$^*$~\cite{thatshoecvpr} & 2016 & Chairs & 297/297 & ~~~~~~~~1   & &  & & \checkmark &  \\ \cline{2-11}
		& \textit{Sketchy}$^*$~\cite{SIGgraph16} & 2016 & Multiple & ~~~75,471/12,500 & ~~~~125   & &  & &  &  \\ \cline{2-11}
		& \textit{QMUL-Handbag}$^*$~\cite{sketchretrievaliccv17} & 2017 & Handbags & ~~~568/568 & ~~~~~~~~1   & &  & &  &  \\ \cline{2-11}
		& \textit{SBIR2017}$^*$~\cite{TIPFGSBIR17} & 2017 & Shoes & ~~912/304 & ~~~~~~~~1   & &\checkmark  & & \checkmark &  \\ \cline{2-11}
		& \textit{QMUL-Shoe-V2}$^*$~\cite{geneCVPRIR19} & 2019 & Shoes & ~~6,730/2,000 & ~~~~~~~~1  & &  & &  &  \\ \cline{2-11}
		& \textit{FG-Xmedia}$^\dagger$~\cite{pengacmmmcross} & 2019 & Birds & 11,788 & ~~~~200  & &  & &  & \checkmark \\ \hline
	\end{tabular}
	\label{table:fgdataset} \\
	\scriptsize{$^*$ For these fine-grained sketch-based image retrieval datasets, normally they have sketch-and-image pairs (\ie, not only images). Thus, we present the numbers of sketches and images separately (the numbers of sketches first). Regarding ``$\sharp$ categories'', we report the number of meta-categories in these datasets.}\\
	\scriptsize{$^\dagger$ Except for text descriptions, \emph{FG-Xmedia} also contains multiple other modalities, \eg, videos and audios.}
\end{table*}

\begin{figure}[t]
\centering
{\includegraphics[width=0.99\columnwidth]{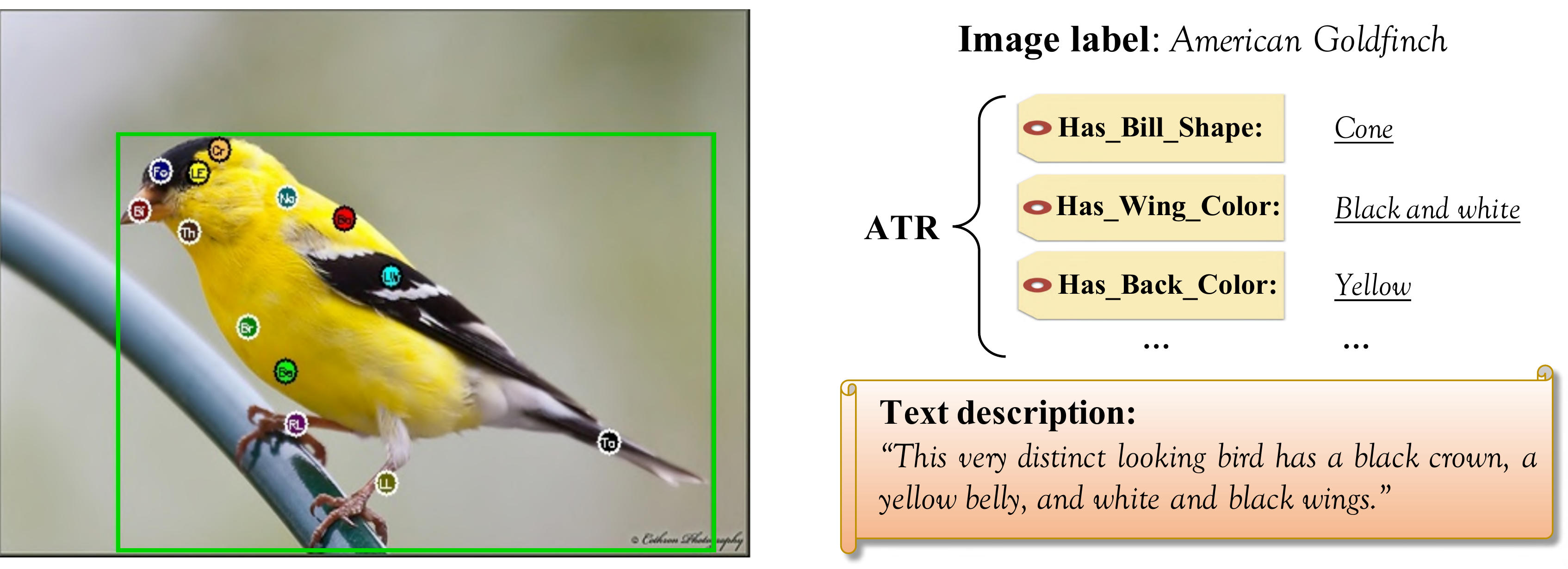}}
\vspace{-0.5em}
\caption{An example image from \emph{CUB200-2011}~\cite{WahCUB200_2011} with multiple different types of annotations \eg, category label, part annotations (\emph{aka} key point locations), object bounding box shown in green, attribute labels (\ie, ``ATR''), and a text description.}
\label{fig:supervisions}
\end{figure}

Representative images from some of these fine-grained benchmark datasets can be found in Figure~\ref{fig:datasetdemo}. In Table~\ref{table:fgdataset}, we summarize the most commonly used image datasets, and indicate their meta-category, the amount of images, the number of categories, their main task, and additional available supervision, \eg, bounding boxes, part annotations, hierarchical labels, attribute labels, and text descriptions (cf. Figure~\ref{fig:supervisions}). These datasets have been one of the most important factors for the considerable progress in the field, not only as a common ground for measuring and comparing performance of competing approaches, but also pushing this field towards increasingly complex, practical, and challenging problems.

\begin{figure*}[t]
\centering
{\includegraphics[width=0.999\textwidth]{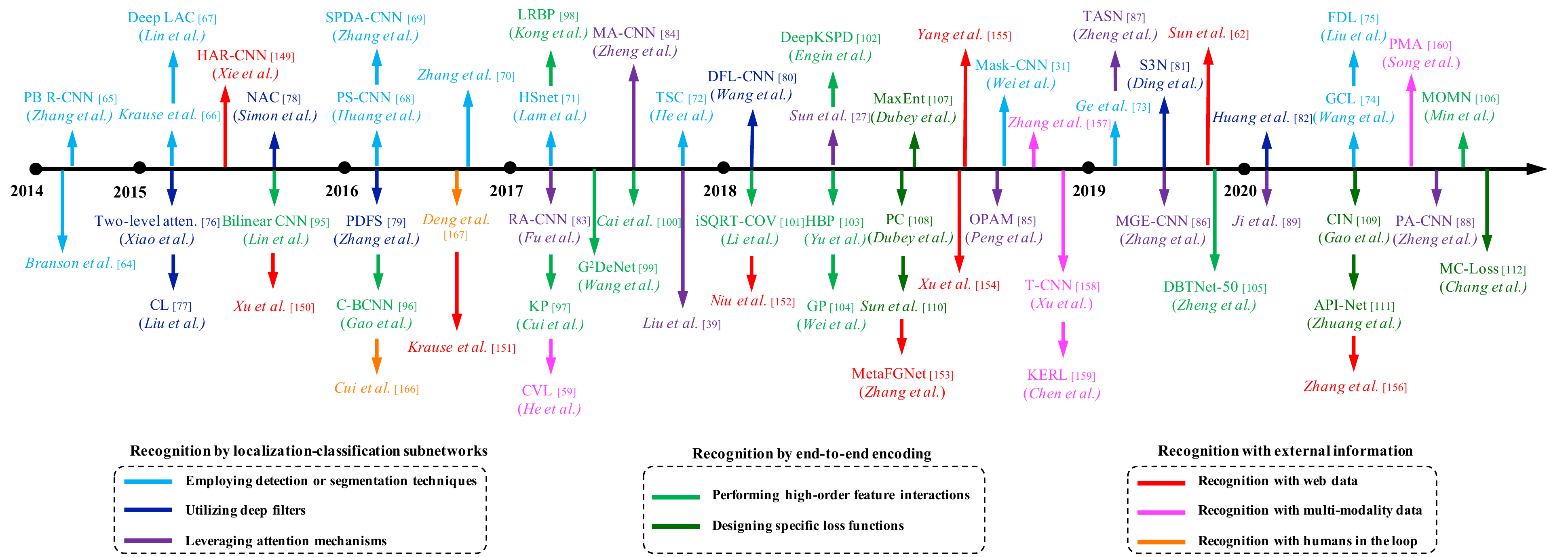}}
\vspace{-20pt}
\caption{Chronological overview of representative deep learning based fine-grained recognition methods which are categorized by different learning approaches. (Best viewed in color.)
}
\label{fig:detailedParadigm}
\end{figure*}

The fine-grained bird classification dataset \emph{CUB200-2011}~\cite{WahCUB200_2011} is one of the most popular fine-grained datasets. 
The majority of FGIA approaches choose it for comparisons with the state-of-the-art. Moreover, continuous contributions are made upon \emph{CUB200-2011} for advanced tasks, \eg, collecting text descriptions of the fine-grained images for multi-modal analysis, cf.~\cite{fgtextcvpr2016,yuxinpengcvpr2017} and Section~\ref{sec:multimodal}.

In recent years, more challenging and practical fine-grained datasets have been proposed, \eg, \emph{iNat2017} containing different species of plants and animals~\cite{inat2017}, and \emph{RPC} for retail products~\cite{rpc}. Novel properties of these datasets include the fact that they are large-scale, have a hierarchical structure, exhibit a domain gap, and form a long-tailed distribution.
These challenges illustrate the practical requirements of FGIA in the real-world and motivate new interesting research challenges (cf. Section~\ref{sec:realsetting}).

Beyond that, a series of fine-grained sketch-based image retrieval datasets, \eg, \emph{QMUL-Shoe}~\cite{thatshoecvpr}, \emph{QMUL-Chair}~\cite{thatshoecvpr}, \emph{QMUL-handbag}~\cite{sketchretrievaliccv17}, \emph{SBIR2014}~\cite{SBIRbmvc14}, \emph{SBIR2017}~\cite{TIPFGSBIR17}, \emph{Sketchy}~\cite{SIGgraph16}, \emph{QMUL-Shoe-V2}~\cite{geneCVPRIR19}, were constructed to further advance the development of fine-grained retrieval, cf. Section~\ref{sec:FGSBIR}. Furthermore, some novel datasets and benchmarks, such as \emph{FG-Xmedia}~\cite{pengacmmmcross}, were constructed to expand fine-grained image retrieval to fine-grained cross-media retrieval.


\section{Fine-Grained Image Recognition}\label{sec:fgrecognition}

Fine-grained image recognition has been by far the most active research area of FGIA in the past decade. Fine-grained recognition aims to discriminate numerous visually similar subordinate categories that belong to the same basic category, such as the fine distinction of animal species~\cite{inat2017}, cars~\cite{cars}, fruits~\cite{vegfru}, aircraft models~\cite{airplanes}, and so on. 
It has been frequently applied in real-world tasks, \eg, ecosystem conservation (recognizing biological species)~\cite{FGICCVworkshop}, intelligent retail systems~\cite{rpc,retailDL}, etc. Recognizing fine-grained categories is difficult due to the challenges of discriminative region localization and fine-grained feature learning. Researchers have attempted to deal with these challenges from diverse perspectives. In this section, we review the main fine-grained recognition approaches since the advent of deep learning.

Broadly, existing fine-grained recognition approaches can be organized into the following three main paradigms:
\begin{itemize}[leftmargin = 2.2em]
\item Recognition by localization-classification subnetworks;
\item Recognition by end-to-end feature encoding;
\item Recognition with external information.
\end{itemize}

Among them, the first two paradigms restrict themselves by only utilizing the supervisions associated with fine-grained images, such as image labels, bounding boxes, part annotations, etc.
To further resolve ambiguous fine-grained problems, there is a body of work that uses additional information such as where and when the image was taken~\cite{presenceICCV2019,chu2019geo}, web images~\cite{xiaxiaoaaai19,niulicvpr18}, or text description~\cite{fgtextcvpr2016,yuxinpengcvpr2017}. In order to present these representative deep learning based fine-grained recognition methods intuitively, we show a chronological overview in Figure~\ref{fig:detailedParadigm} by organizing them into the three aforementioned paradigms.

For performance evaluation, when the test set is balanced (\ie, there is a similar number test examples from each class), the most commonly used metric in fine-grained recognition is classification \emph{accuracy} across all subordinate categories of the datasets. It is defined as
\begin{equation}
{\rm Accuracy} = \frac{|I_{\rm correct}|}{|I_{\rm total}|}\,,
\end{equation}
where $|I_{\rm total}|$ represents the number of images across all sub-categories in the test set and $|I_{\rm correct}|$ represents the number of images which are correctly categorized by the model.

\subsection{Recognition by Localization-Classification Subnetworks}\label{sec:loc-cls}

\begin{figure}[t!]
\centering
	{\includegraphics[width=0.95\columnwidth]{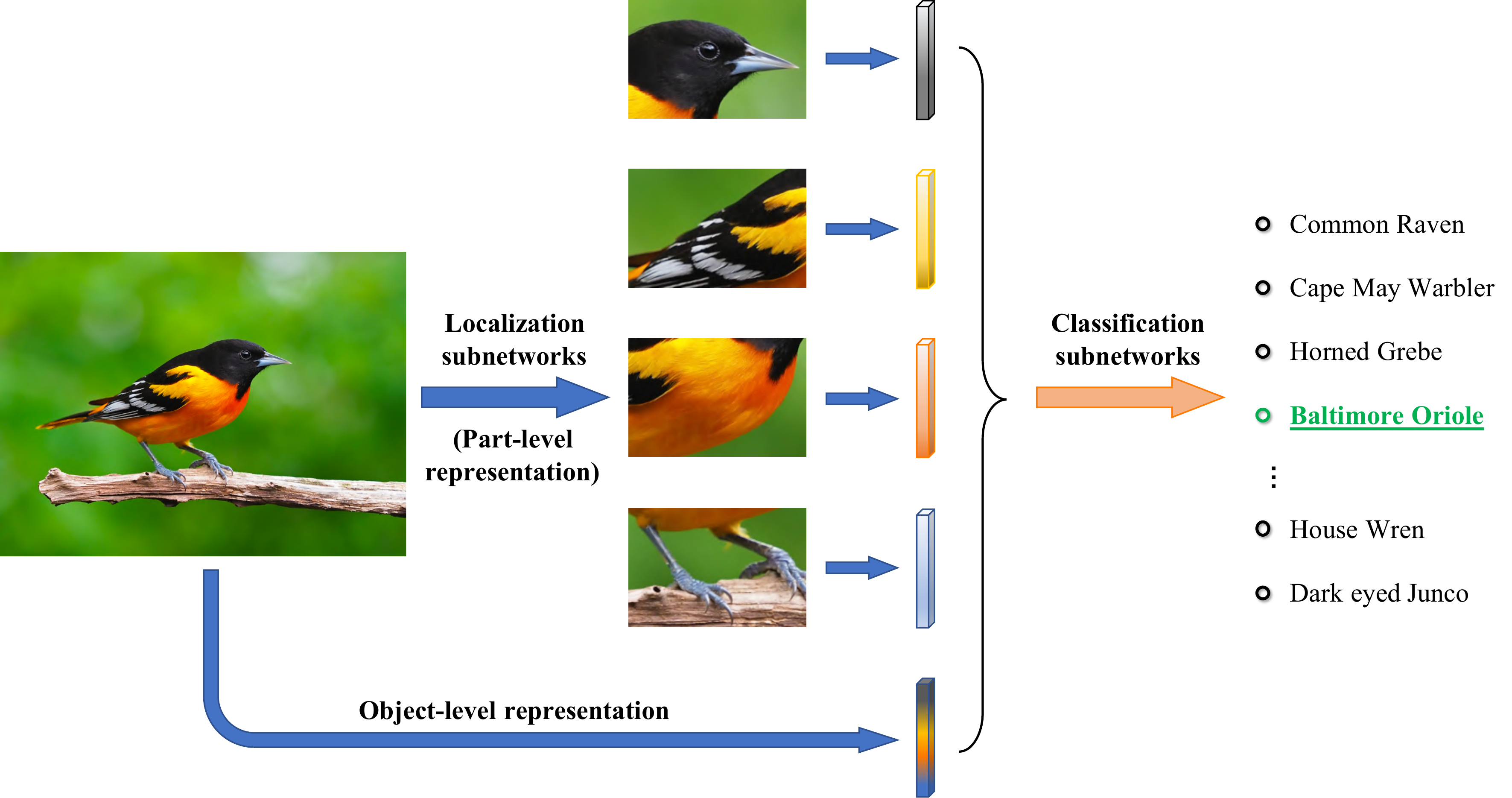}}
	\vspace{-0.5em}
\caption{Illustration of the high-level pipeline of the fine-grained recognition by localization-classification subnetworks paradigm.}
\label{fig:loc_cls}
\end{figure}

Researchers have attempted to create models that capture the discriminative semantic parts of fine-grained objects and then construct a mid-level representation corresponding to these parts for the final classification, cf. Figure~\ref{fig:loc_cls}.
More specifically, a localization subnetwork is designed for locating  key parts, and then the corresponding part-level (local) feature vectors are obtained. 
This is usually combined with object-level (global) image features for representing fine-grained objects. This is followed by a classification subnetwork which performs recognition. The framework of such two collaborative subnetworks forms the first paradigm, \ie, fine-grained recognition with \emph{localization-classification subnetworks}. The motivation for these models is to first find the corresponding parts and then compare their appearance. Concretely, it is desirable to capture semantic parts (\eg, heads and torsos) that are shared across fine-grained categories and for discovering the subtle differences between these part representations.

Existing methods in this paradigm can be divided into four broad types: 1) employing detection or segmentation techniques, 2) utilizing deep filters, 3) leveraging attention mechanisms, and 4) other methods.

\begin{table*}[h!]
\scriptsize
\caption{Comparative fine-grained recognition results of two learning paradigms (cf. Section~\ref{sec:loc-cls} and Section~\ref{sec:end2end}) on the fine-grained benchmark datasets, \ie, Birds (\emph{CUB200-2011}~\cite{WahCUB200_2011}), Dogs (\emph{Stanford Dogs}~\cite{Khosla11stanforddogs}), Cars (\emph{Stanford Cars}~\cite{cars}), and Aircrafts (\emph{FGVC Aircraft}~\cite{airplanes}). Note that, ``Train anno.'' and ``Test anno.'' mean which supervised signals used in the training and test phases, respectively. The symbol ``--'' means the results are unavailable.}
\vspace{-1em}
\label{table:bigtablerecog1}
\setlength{\tabcolsep}{5.3pt}
\begin{tabular}{|c|l|c|c|c|c|c|c||c|c|c|c|}
\hline
\multicolumn{3}{|c|}{\multirow{2}{*}{Methods}}                                                                                                                                                                                        & \multirow{2}{*}{Published in} & \multirow{2}{*}{Train anno.} & \multirow{2}{*}{Test anno.} & \multirow{2}{*}{Backbones} & \multirow{2}{*}{Img. resolution} & \multicolumn{4}{c|}{Accuracy}        \\ \cline{9-12} 
\multicolumn{3}{|c|}{}                                                                                                                                                                                                                &                               &                              &                             &                            &                                  & \textit{Birds}  & \textit{Dogs}   & \textit{Cars}   & \textit{Aircrafts} \\ \hline\hline
\multirow{35}{*}{\rotatebox{90}{\tiny Fine-grained recognition by localization-classification subnetworks}} & \multirow{14}{*}{\rotatebox{90}{\tiny Employing detection or segmentation techniques}} & Branson~\etal~\cite{branson2014bird}                & BMVC 2014                     & BBox+Parts                   &                        & CaffeNet                   & $224\times 224$                  & 75.7\% & --     & --     & --        \\ \cline{3-12} 
& & PB R-CNN~\cite{Ning14ECCV}                 & ECCV 2014                     & BBox+Parts                   & BBox                        & Alex-Net                   & $224\times 224$                  & 76.4\% & --     & --     & --        \\ \cline{3-12} 
                                                                                                      &                                                                                  & Krause \etal~\cite{krausecvpr15}           & CVPR 2015                     & BBox                         &                             & CaffeNet                   & $224\times 224$                  & 82.0\% & --     & 92.6\% & --        \\ \cline{3-12} 
                                                                                                      &                                                                                  & Deep LAC~\cite{Di15CVPR}                   & CVPR 2015                     & BBox+Parts                   & BBox                        & Alex-Net                   & $227\times 227$                  & 80.3\% & --     & --     & --        \\ \cline{3-12} 
                                                                                                      &                                                                                  & PS-CNN~\cite{pscnncvpr16}                  & CVPR 2016                     & BBox+Parts                   & BBox                        & CaffeNet                   & $227\times 227$                  & 76.6\% & --     & --     & --        \\ \cline{3-12} 
                                                                                                      &                                                                                  & SPDA-CNN~\cite{spdacvpr16cnn}              & CVPR 2016                     & BBox+Parts                   & BBox                        & CaffeNet                   & Longer side to 800px             & 81.0\% & --     & --     & --        \\ \cline{3-12} 
                                                                                                      &                                                                                  & SPDA-CNN~\cite{spdacvpr16cnn}              & CVPR 2016                     & BBox+Parts                   & BBox                        & VGG-16                     & Longer side to 800px             & 85.1\% & --     & --     & --        \\ \cline{3-12} 
                                                                                                      &                                                                                  & Zhang \etal~\cite{zhangyu16tipweakly}      & IEEE TIP 2016                 &                              &                             & Alex-Net                   & $224\times 224$                  & 78.9\% & 80.4\% & --     & --        \\ \cline{3-12} 
                                                                                                      &                                                                                  & HSnet~\cite{hsnetcvpr17}                   & CVPR 2017                     & Parts                        &                             & GoogLeNet                  & $224\times 224$                  & 87.5\% & --     & 93.9\% & --        \\ \cline{3-12} 
                                                                                                      &                                                                                  & TSC~\cite{XHeAAAI17FG}                     & AAAI 2017                     &                              &                             & VGG-16                     & Not given                        & 84.7\% & --     & --     & --        \\ \cline{3-12} 
                                                                                                      &                                                                                  & Mask-CNN~\cite{maskcnnPR}                  & PR 2018                       & Parts                        &                             & VGG-16                     & $448\times 448$                  & 85.7\% & --     & --     & --        \\ \cline{3-12} 
                                                                                                      &                                                                                  & Ge \etal~\cite{yizhoucvpr19}              & CVPR 2019                     &                              &                             & GoogLeNet+BN               & Shorter side to 800px            & 90.3\% & 93.9\% & --     & --        \\ \cline{3-12} 
                                                                                                      &                                                                                  & GCL~\cite{graphAAAI20}                     & AAAI 2020                     &                              &                             & ResNet-50+BN               & $448\times 448$                  & 88.3\% & --     & 94.0\% & 93.2\%    \\ \cline{3-12} 
                                                                                                      &                                                                                  & FDL~\cite{FDLAAAI2020}                     & AAAI 2020                     &                              &                             & ResNet-50                  & $448\times 448$                  & 88.6\% & 85.0\% & 94.3\% & 93.4\%    \\ \cline{2-12} 
                                                                                                      & \multirow{7}{*}{\rotatebox{90}{\tiny Utilizing deep filters}}                          & Two-level atten.~\cite{xiaoCVPR15twolevel} & CVPR 2015                     &                              &                             & VGG-16                     & Not given                        & 77.9\% & --     & --     & --        \\ \cline{3-12} 
                                                                                                      &                                                                                  & CL~\cite{treasureLQ2015}                   & CVPR 2015                     &                              &                             & Alex-Net                   & $448\times 448$                  & 73.5\% & --     & --     & --        \\ \cline{3-12} 
                                                                                                      &                                                                                  & NAC~\cite{nacICCV2015}                     & ICCV 2015                     &                              &                             & VGG-19                     & Not given                        & 81.0\% & --     & --     & --        \\ \cline{3-12} 
                                                                                                      &                                                                                  & PDFS~\cite{pickingcvpr2016}                & CVPR 2016                     &                              &                             & VGG-16                     & Not given                        & 84.5\% & 72.0\% & --     & --        \\ \cline{3-12} 
                                                                                                      &                                                                                  & DFL-CNN~\cite{filterbankCVPR2018}          & CVPR 2018                     &                              &                             & VGG-16                     & $448\times 448$                  & 86.7\% & --     & 93.8\% & 92.0\%    \\ \cline{3-12} 
                                                                                                      &                                                                                  & S3N~\cite{s3niccv2019}                     & ICCV 2019                     &                              &                             & ResNet-50                  & $448\times 448$                  & 88.5\% & --     & 94.7\% & 92.8\%    \\ \cline{3-12} 
                                                                                                      &                                                                                  & Huang \etal~\cite{huangCVPR2020inter}      & CVPR 2020                     &                              &                             & ResNet-101                 & Shorter side to 448px            & 87.3\% & --     & --     & --        \\ \cline{2-12} 
                                                                                                      & \multirow{9}{*}{\rotatebox{90}{\tiny Attention mechanisms}}                            & RA-CNN~\cite{RACNN}                        & CVPR 2017                     &                              &                             & VGG-19                     & $448\times 448$                  & 85.3\% & 87.3\% & 92.5\% & --        \\ \cline{3-12} 
                                                                                                      &                                                                                  & MA-CNN~\cite{MACNN}                        & ICCV 2017                     &                              &                             & VGG-19                     & $448\times 448$                  & 86.5\% & --     & 92.8\% & 89.9\%    \\ \cline{3-12} 
                                                                                                      &                                                                                  & Liu \etal~\cite{yuanqingaaai17local}       & AAAI 2017                     & Parts+Attr.                  &                             & ResNet-50                  & $448\times 448$                  & 85.4\% & --     & --     & --        \\ \cline{3-12} 
                                                                                                      &                                                                                  & Sun \etal~\cite{MAMCeccv}                  & ECCV 2018                     &                              &                             & ResNet-50                  & $448\times 448$                  & 86.5\% & 84.8\% & 93.0\% & --        \\ \cline{3-12} 
                                                                                                      &                                                                                  & OPAM~\cite{pengTIPattfg}                   & IEEE TIP 2018                 &                              &                             & VGG-16                     & Not given                        & 85.8\% & --     & 92.2\% & --        \\ \cline{3-12} 
                                                                                                      &                                                                                  & MGE-CNN~\cite{mixureexperICCV19}           & ICCV 2019                     &                              &                             & ResNet-50                  & $448\times 448$                  & 88.5\% & --     & 93.9\% & --        \\ \cline{3-12} 
                                                                                                      &                                                                                  & TASN~\cite{triliattention19}               & CVPR 2019                     &                              &                             & ResNet-50                  & $224\times 224$                  & 87.9\% & --     & 93.8\% & --        \\ \cline{3-12} 
                                                                                                      &                                                                                  & PA-CNN~\cite{rphheliangtip2020}            & IEEE TIP 2020                 &                              &                             & VGG-19                     & $448\times 448$                  & 87.8\% & --     & 93.3\% & 91.0\%    \\ \cline{3-12} 
                                                                                                      &                                                                                  & Ji \etal~\cite{attTreefgvc20}              & CVPR 2020                     &                              &                             & ResNet-50                  & $448\times 448$                  & 88.1\% & --     & 94.6\% & 92.4\%    \\ \cline{2-12} 
                                                                                                      & \multirow{5}{*}{\rotatebox{90}{\tiny Others}}                                          & STN~\cite{STN}                             & NeurIPS 2015                  &                              &                             & GoogLeNet+BN               & $448\times 448$                  & 84.1\% & --     & --     & --        \\ \cline{3-12} 
                                                                                                      &                                                                                  & BoT~\cite{daviscvpr16triplet}              & CVPR 2016                     &                              &                             & Alex-Net                   & Not given                        & --     & --     & 92.5\% & 88.4\%    \\ \cline{3-12} 
                                                                                                      &                                                                                  & NTS-Net~\cite{navigateECCV18}              & ECCV 2018                     &                              &                             & ResNet-50                  & $448\times 448$                  & 87.5\% & --     & 93.9\% & 91.4\%    \\ \cline{3-12} 
                                                                                                      &                                                                                  & M2DRL~\cite{xiangtengIJCVfg}               & IJCV 2019                     &                              &                             & VGG-16                     & $448\times 448$                  & 87.2\% & --     & 93.3\% & --        \\ \cline{3-12} 
                                                                                                      &                                                                                  & DF-GMM~\cite{gaussCVPR20fg}                & CVPR 2020                     &                              &                             & ResNet-50                  & $448\times 448$                  & 88.8\% & --     & 94.8\% & 93.8\%    \\ \hline\hline
\multirow{25}{*}{\rotatebox{90}{\tiny Fine-grained recognition by end-to-end feature encoding}}             & \multirow{11}{*}{\rotatebox{90}{\tiny High-order feature interactions}}                & Bilinear CNN~\cite{TsungYu15ICCV}          & ICCV 2015                     &                              &                             & VGG-16+VGG-M             & $448\times 448$                  & 84.1\% & --     & 91.3\% & 84.1\%    \\ \cline{3-12} 
                                                                                                      &                                                                                  & C-BCNN~\cite{compactBCNN}                  & CVPR 2016                     &                              &                             & VGG-16                     & $448\times 448$                  & 84.3\% & --     & 91.2\% & 84.1\%    \\ \cline{3-12} 
                                                                                                      &                                                                                  & KP~\cite{yinkernel17}                      & CVPR 2017                     &                              &                             & VGG-16                     & $224\times 224$                  & 86.2\% & --     & 92.4\% & 86.9\%    \\ \cline{3-12} 
                                                                                                      &                                                                                  & LRBP~\cite{lowrankBCNN}                    & CVPR 2017                     &                              &                             & VGG-16                     & $224\times 224$                  & 84.2\% & --     & 90.0\% & 87.3\%    \\ 
\cline{3-12} 
                                                                                                      &                                                                                  & G$^2$DeNet~\cite{g2denetcvpr}                    & CVPR 2017                     &                              &                             & VGG-16                     & Longer side to 200px                  & 87.1\% & --     & 92.5\% & 89.0\% \\\cline{3-12} 
                                                                                                      &                                                                                  & Cai~\etal~\cite{highorleizhangiccv17}      & ICCV 2017                     &                              &                             & VGG-16                     & $448\times 448$                  & 85.3\% & --     & 91.7\% & 88.3\%    \\ \cline{3-12} 
                                                                                                      &                                                                                  & iSQRT-COV~\cite{matsrnormcvpr18}           & CVPR 2018                     &                              &                             & ResNet-101                 & $224\times 224$                  & 88.7\% & --     & 93.3\% & 91.4\%    \\ \cline{3-12} 
                                                                                                      &                                                                                  & DeepKSPD~\cite{deepKSPDeccv18}             & ECCV 2018                     &                              &                             & VGG-16                     & $448\times 448$                  & 86.5\% & --     & 93.2\% & 91.0\%    \\ \cline{3-12} 
                                                                                                      &                                                                                  & HBP~\cite{hbp18eccv}                       & ECCV 2018                     &                              &                             & VGG-16                     & $448\times 448$                  & 87.1\% & --     & 93.7\% & 90.3\%    \\ \cline{3-12} 
                                                                                                      &                                                                                  & GP~\cite{grassmannpool}                    & ECCV 2018                     &                              &                             & VGG-16                     & $448\times 448$                  & 85.8\% & --     & 92.8\% & 89.8\%    \\ \cline{3-12} 
                                                                                                      &                                                                                  & DBTNet-50~\cite{deepbcnnNIPS19}            & NeurIPS 2019                  &                              &                             & VGG-16                     & $448\times 448$                  & 87.5\% & --     & 94.1\% & 91.2\%    \\ \cline{3-12} 
                                                                                                      &                                                                                  & MOMN~\cite{MOMNtip2020}                    & IEEE TIP 2020                 &                              &                             & VGG-16                     & $448\times 448$                  & 87.3\% & --     & 92.8\% & 90.4\%    \\ \cline{2-12} 
                                                                                                      & \multirow{9}{*}{\rotatebox{90}{\tiny Specific loss functions}}                         & MaxEnt~\cite{maximumEntro}                 & NeurIPS 2018                  &                              &                             & VGG-16                     & $224\times 224$                  & 77.0\% & 65.4\% & 83.9\% & 78.1\%    \\ \cline{3-12} 
                                                                                                      &                                                                                  & MaxEnt~\cite{maximumEntro}                 & NeurIPS 2018                  &                              &                             & Bilinear CNN               & $224\times 224$                  & 85.3\% & 83.2\% & 92.8\% & 86.1\%    \\ \cline{3-12} 
                                                                                                      &                                                                                  & PC~\cite{pariECCV18}                       & ECCV 2018                     &                              &                             & Bilinear CNN               & $224\times 224$                  & 85.6\% & 83.0\% & 92.4\% & 85.7\%    \\ \cline{3-12} 
                                                                                                      &                                                                                  & Sun \etal~\cite{MAMCeccv}                  & ECCV 2018                     &                              &                             & ResNet-50                  & $448\times 448$                  & 86.5\% & 84.8\% & 93.0\% & --        \\ \cline{3-12} 
                                                                                                      &                                                                                  & CIN~\cite{CINAAAI2020}                     & AAAI 2020                     &                              &                             & ResNet-101                 & $448\times 448$                  & 88.1\% & --     & 94.5\% & 92.8\%    \\ \cline{3-12} 
                                                                                                      &                                                                                  & Sun~\etal~\cite{lingshaoaaai20ac}          & AAAI 2020                     &                              &                             & ResNet-50                  & $448\times 448$                  & 88.6\% & 87.7\% & 94.9\% & 93.5\%    \\ \cline{3-12} 
                                                                                                      &                                                                                  & API-Net~\cite{pairinterQiaoAAAI20}         & AAAI 2020                     &                              &                             & ResNet-50                  & $448\times 448$                  & 87.7\% & 88.3\% & 94.8\% & 93.0\%    \\ \cline{3-12} 
                                                                                                      &                                                                                  & API-Net~\cite{pairinterQiaoAAAI20}         & AAAI 2020                     &                              &                             & DenseNet-161               & $448\times 448$                  & 90.0\% & 89.4\% & 95.3\% & 93.9\%    \\ \cline{3-12} 
                                                                                                      &                                                                                  & MC-Loss~\cite{mclossTIP20}                 & IEEE TIP 2020                 &                              &                             & Bilinear CNN               & $448\times 448$                  & 86.4\% & --     & 94.4\% & 92.9\%    \\ \cline{2-12} 
                                                                                                      & \multirow{5}{*}{\rotatebox{90}{\tiny Others}}                                          & BGL~\cite{BGLCVPR2016}                     & CVPR 2016                     &                              &                             & VGG-16                     & $224\times 224$                  & 75.9\% & --     & 86.0\% & --        \\ \cline{3-12} 
                                                                                                      &                                                                                  & BGL~\cite{BGLCVPR2016}                     & CVPR 2016                     & BBox                         &                             & VGG-16                     & $224\times 224$                  & 80.4\% & --     & 90.5\% & --        \\ \cline{3-12} 
                                                                                                      &                                                                                  & DCL~\cite{destructioncvpr19}               & CVPR 2019                     &                              &                             & ResNet-50                  & $448\times 448$                  & 87.8\% & --     & 94.5\% & 93.0\%    \\ \cline{3-12} 
                                                                                                      &                                                                                  & Cross-X~\cite{crossXiccv19}                & ICCV 2019                     &                              &                             & ResNet-50                  & $448\times 448$                  & 87.7\% & 88.9\% & 94.6\% & 92.6\%    \\ \cline{3-12} 
                                                                                                      &                                                                                  & PMG~\cite{FGjigsaw20}                      & ECCV 2020                     &                              &                             & ResNet-50                  & $448\times 448$                  & 89.6\% & --     & 95.1\% & 93.4\%    \\ \hline

\end{tabular}
\end{table*}

\subsubsection{Employing Detection or Segmentation Techniques}\label{sec:det-seg}

It is straightforward to employ detection or segmentation techniques~\cite{fasterrcnn,fcn15,rcnnCVPR14} to locate key image regions corresponding to fine-grained object parts, \eg, bird heads, bird tails, car lights, dog ears, dog torsos, etc. Thanks to localization information, \ie, part-level bounding boxes or segmentation masks, the model can obtain more discriminative mid-level (part-level) representations w.r.t. these parts. Thus, it could further enhance the learning capability of the classification subnetwork, thus significantly boost the final recognition accuracy.

Earlier works in this paradigm made use of additional dense part annotations (\emph{aka} key point localization, cf. Figure~\ref{fig:supervisions} on the left) to locate semantic key parts of objects. For example, Branson~\etal~\cite{branson2014bird} proposed to use groups of detected part keypoints to compute multiple warped image regions and further obtained the corresponding part-level features by pose normalization. In the same period, Zhang~\etal~\cite{Ning14ECCV} first generated part-level bounding boxes based on ground truth part annotations, and then trained a R-CNN~\cite{rcnnCVPR14} model to perform part detection. Di~\etal~\cite{Di15CVPR} further proposed a Valve Linkage Function, which not only connected all subnetworks, but also refined localization according to the \emph{part alignment} results. In order to integrate both semantic part detection and abstraction, SPDA-CNN~\cite{spdacvpr16cnn} designed a top-down method to generate \emph{part-level} proposals by inheriting prior geometric constraints and then used a Faster R-CNN~\cite{fasterrcnn} to return part localization predictions. 
Other approaches made use of segmentation information.
PS-CNN~\cite{pscnncvpr16} and Mask-CNN~\cite{maskcnnPR} employed segmentation models to get part/object masks to aid part/object localization. Compared with detection techniques, segmentation can result in more accurate part localization~\cite{maskcnnPR} as segmentation focuses on the finer pixel-level targets, instead of just coarse bounding boxes.

However, employing traditional detectors or segmentation models requires dense part annotations for training, which is labor-intensive and would limit both scalability and practicality of real-world fine-grained applications. Therefore, it is desirable to accurately locate fine-grained parts by only using image level labels~\cite{zhangyu16tipweakly,XHeAAAI17FG,yizhoucvpr19,graphAAAI20,FDLAAAI2020}. These set of approaches are referred to as ``weakly-supervised'' as they only use image level labels. It is interesting to note that since 2016 there is an apparent trend in developing fine-grained methods in this weakly-supervised setting, rather than the strong-supervised setting (\ie, using part annotations and bounding boxes), cf. Table~\ref{table:bigtablerecog1}.

Recognition methods in the weakly-supervised localization based classification setting always rely on unsupervised approaches to obtain semantic groups which correspond to object parts. Specifically, Zhang~\etal~\cite{zhangyu16tipweakly} adopted the spatial pyramid strategy~\cite{spjponce06} to generate part proposals from object proposals. Then, by using a clustering approach, they generated part proposal prototype clusters and further selected useful clusters to get discriminative part-level features. 
Co-segmentation~\cite{guillauminIJCV2014} based methods are also commonly used in this weakly supervised case. 
One approach is to use co-segmentation to obtain object masks without supervision, and then perform heuristic strategies, \eg, part constraints~\cite{XHeAAAI17FG} or part alignment~\cite{krausecvpr15}, to locate fine-grained parts.

It is worth noting that the majority of previous work overlooks the internal semantic correlation among discriminative part-level features. Concretely, the aforementioned methods pick out the discriminative regions independently and utilize their features directly, while neglecting the fact that an object's features are mutually semantic correlated and region groups can be more discriminative. Therefore, very recently, some methods attempt to jointly learn the interdependencies among part-level features to obtain more universal and powerful fine-grained image representations. By performing different feature fusion strategies (\eg, LSTMs~\cite{hsnetcvpr17,yizhoucvpr19}, graphs~\cite{graphAAAI20}, or knowledge distilling~\cite{FDLAAAI2020}) these joint part feature learning methods yield significantly higher recognition accuracy over previous independent part feature learning methods.

\subsubsection{Utilizing Deep Filters}

In deep convolutional neural networks (CNNs), deep filters (\ie, CNN filters) refer to the learned weights of the convolution layers~\cite{natureDL}. The responses/activations from these deep filters can be viewed as localized descriptors. The deep descriptors have the following properties~\cite{DDTPR}: 1) Locality: they describe and correspond to local image regions w.r.t. the whole input image and 2) Spatiality: they are also able to encode spatial information. 
As works started exploring the use of CNNs in computer vision, researchers gradually discovered that intermediate CNN outputs (\eg, local deep descriptors) could be linked to semantic parts of common objects~\cite{visualeccv14}. Therefore, the fine-grained community attempted to employ these filter outputs as part detectors~\cite{xiaoCVPR15twolevel,treasureLQ2015,nacICCV2015,pickingcvpr2016,filterbankCVPR2018,s3niccv2019,huangCVPR2020inter}, and thus rely on them to conduct localization-classification fine-grained recognition. 
One of the main advantages here is that this does not require any part-level annotations. 

Xiao \etal~\cite{xiaoCVPR15twolevel} performed spectral clustering~\cite{PRML} on these deep filters to form groups and then used the filter groups to serve as part detectors. Similarly, NAC~\cite{nacICCV2015} exploits the channels of a CNN as part detectors. Liu~\etal~\cite{treasureLQ2015} developed a cross-layer pooling method by aggregating the part-based deep descriptors using activations from two successive convolutional layers as guidance. PDFS~\cite{pickingcvpr2016} was proposed to select deep filters corresponding to parts and then iteratively update the learned ``detectors'' which resulted in the discovery of discriminative and consistent part-based image regions. For these aforementioned methods, after obtaining detected parts using deep filters from pre-trained classification CNNs, they typically trained off-line classifiers, \eg, SVMs or decision trees~\cite{PRML}, using the part-based feature vectors to conduct the final recognition task.

To facilitate the learning of  both part detection and part-based classification, unified end-to-end trained fine-grained models~\cite{filterbankCVPR2018,s3niccv2019,huangCVPR2020inter} were developed. 
As a result, significant recognition improvements were observed, cf. Table~\ref{table:bigtablerecog1}. 
Wang \etal~\cite{filterbankCVPR2018} utilized an additional learnable $1\times 1$ convolutional filter as a small patch (\ie, part) detector. 
This was followed by a global max-pooling to keep the highest activations w.r.t. that filter for the final classification. 
Later, based on class response maps~\cite{boleicvpr16loc}, S3N~\cite{s3niccv2019} leveraged the class peak responses, \ie, local maximums, as the basis of part localization. 
Similar to~\cite{hsnetcvpr17,yizhoucvpr19,graphAAAI20}, S3N also considers part correlations in a mutual part learning way.

\subsubsection{Leveraging Attention Mechanisms}

Even though the previous localization-classification fine-grained methods have shown strong classification performance, one of their major drawbacks is that they require meaningful definitions of the object parts. In many applications however, it may be hard to represent or even define common parts of some object classes, \eg, non-structured objects like food dishes~\cite{food101} or flowers with repeating parts~\cite{Flowers08}. Compared to these localization-classification methods, a more natural solution of finding parts is to leverage attention mechanisms~\cite{pami98attention} as sub-modules. This enables CNNs to attend to loosely defined regions for fine-grained objects and as a result have emerged as a promising direction.

It is common knowledge that attention plays an important role in human perception~\cite{pami98attention,natureneuro02}. Humans exploit a sequence of partial glimpses and selectively focus on salient parts of an object or a scene in order to better capture visual structure~\cite{nipshiton2010}. 
Inspired by this, Fu~\etal~and Zheng~\etal~\cite{RACNN,MACNN} were the first to incorporate attention processing to improve the fine-grained recognition accuracy of CNNs. Specifically, RA-CNN~\cite{RACNN} uses a recurrent visual attention model to select a sequence of attention regions (corresponding to object ``parts''\footnote{Note that here ``parts'' refers to the loosely defined attention regions for fine-grained objects, which is different from the clearly defined object parts from manual annotations, cf. Section~\ref{sec:det-seg}.}). RA-CNN iteratively generates region attention maps in a coarse to fine fashion by taking previous predictions as a reference. MA-CNN~\cite{MACNN} is equipped with a multi-attention CNN, and can return multiple region attentions in parallel. Subsequently, Peng~\etal~\cite{pengTIPattfg} and Zheng~\etal~\cite{rphheliangtip2020} proposed multi-level attention models to obtain hierarchical attention information (\ie, both object- and part-level). He~\etal~\cite{tcsvthexiangt} applied multi-level attention to localize multiple discriminative regions simultaneously for each image via an $n$-pathway end-to-end discriminative localization network that simultaneously localizes discriminative regions and encodes their features.
This multi-level attention can result in diverse and complementary information compared to the aforementioned single-level attention methods. 
Sun~\etal~\cite{MAMCeccv} incorporated channel attentions~\cite{SEnet} and metric learning~\cite{siamesenips93} to enforce the correlations among different attended regions. Zheng~\etal~\cite{triliattention19} developed a trilinear attention sampling network to learn fine-grained details from hundreds of part proposals and efficiently distill the learned features into a single CNN. 
Recently, Ji~\etal~\cite{attTreefgvc20} presented an attention based convolutional binary neural tree, which incorporates attention mechanisms with a tree structure to facilitate  coarse-to-fine hierarchical fine-grained feature learning. Although the attention mechanism achieves strong accuracy in fine-grained recognition, it tends to overfit in the case of small-scale data.

\subsubsection{Other Methods}

Many other approaches in the localization-classification paradigm have also been proposed for fine-grained recognition.
Spatial Transformer Networks (STN)~\cite{STN} were originally introduced to explicitly perform spatial transformations in an end-to-end learnable way. 
They can also be equipped with multiple transformers in parallel to conduct fine-grained recognition. 
Each transformer in an STN can correspond to a part detector with spatial transformation capabilities. Later, Wang~\etal~\cite{daviscvpr16triplet} developed a triplet of patches with geometric constraints as a template to automatically mine discriminative triplets and then generated mid-level representations for classification with the mined triplets. 
In addition, other methods have achieved better accuracy by introducing feedback mechanisms. Specifically, NTS-Net~\cite{navigateECCV18} employs a multi-agent cooperative learning scheme to address the core problem of fine-grained recognition, \ie, accurately identifying informative regions in an image. M2DRL~\cite{ijcaistackRL,xiangtengIJCVfg} was the first to utilize deep reinforcement learning~\cite{RLpaper} at both the object- and part-level to capture multi-granularity discriminative localization and multi-scale representations using their tailored reward functions. 
Inspired by low-rank mechanisms in natural language processing~\cite{acllowrank}, Wang~\etal~\cite{gaussCVPR20fg} proposed the DF-GMM framework to alleviate the region diffusion problem in high-level feature maps for fine-grained part localization. 
DF-GMM first selects discriminative regions from the high-level feature maps by constructing low-rank bases, and then applies spatial information of the low-rank bases to reconstruct low-rank feature maps. 
Part correlations can also be modeled by reorganization processing, which brings accuracy improvements.

\subsection{Recognition by End-to-End Feature Encoding}\label{sec:end2end}

The second learning paradigm of fine-grained recognition is \emph{end-to-end feature encoding}. As with other vision tasks, feature learning also plays a fundamental role in fine-grained recognition.
Since the differences between sub-categories are typically very subtle and local, capturing global semantic information using only fully connected layers limits the representation capacity of a fine-grained model, and hence restricts further improvements in final recognition performance. Therefore, methods have been developed that aim to learn a unified, yet discriminative, image representation for modeling subtle differences between fine-grained categories in the following ways: 1) by performing high-order feature interactions, 2) by designing novel loss functions, and 3) through other means.

\subsubsection{Performing High-Order Feature Interactions}

\begin{figure}[t!]
\centering
	{\includegraphics[width=0.7\columnwidth]{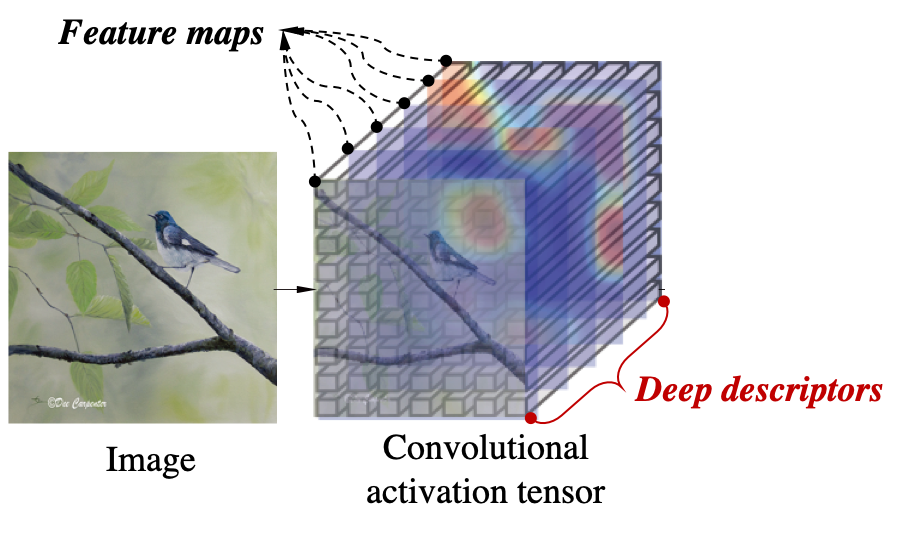}}
	\vspace{-1.5em}
\caption{Illustration of feature maps and deep descriptors in CNNs.}
\label{fig:desc}
\end{figure}

Feature learning plays a crucial role in almost all vision tasks such as retrieval, detection, tracking, etc. The success of deep convolutional networks is mainly due to the learned discriminative deep features. In the initial era of deep learning, the features (\ie, activations) of fully connected layers were commonly used as image representations. Later, the feature maps of deeper convolutional layers were discovered to contain mid- and high-level information, \eg, object parts or complete objects~\cite{iccv2011midlevel}, which led to the widespread use of convolutional features/descriptors~\cite{treasureLQ2015,discrizhongwen15cvpr}, cf. Figure~\ref{fig:desc}. Additionally, applying encoding techniques for these local convolutional descriptors has resulted in significant improvements compared with  fully-connected outputs~\cite{dlbanksav15,discrizhongwen15cvpr,dsp15arxiv}. To some extent, these improvements in encoding techniques come from the \emph{higher-order} statistics encoded in the final features. Particularly for fine-grained recognition, where the need for end-to-end modeling of higher-order statistics became evident when the Fisher Vector encodings of SIFT features outperformed a fine-tuned AlexNet in several fine-grained tasks~\cite{revisitPRL}. 

The covariance matrix based representation~\cite{matrixlei15,qilongtpamicov} is a representative higher-order (\ie, second-order) feature interaction technique, which has been used in computer vision and machine learning. 
Let $\bm{V}_{d\times n}=\left[\bm{v}_1, \bm{v}_2, \ldots, \bm{v}_n \right]$ denote a data matrix, in which each column contains a local descriptor $\bm{v}_i \in \mathcal{R}^d$, extracted from an image. The corresponding $d\times d$ sample covariance matrix over $\bm{V}$ is denoted as $\bm{\Sigma} = \bar{\bm{V}}\bar{\bm{V}}^\top$ (or simply ${\bm{V}}{\bm{V}}^\top$), where $\bar{\bm{V}}$ denotes the centered $\bm{V}$. Originally, this covariance matrix is proposed as a region descriptor, \eg, characterizing the covariance of the color intensities of pixels in an image patch. In recent years, it has been used as a promising second-order pooled image representation for visual recognition.

By integrating the covariance matrix based representation with deep descriptors, a series of methods showed promising accuracy in fine-grained recognition in the past few years. The most representative method among them is Bilinear CNNs~\cite{TsungYu15ICCV,bcnnTPAMI}, which represents an image as a pooled outer product of features derived from two deep CNNs, and thus encodes second-order statistics of convolutional activations, resulting in clear improvements in fine-grained recognition. This outer product essentially leads to a covariance matrix (in the form of $\bm{V}\bm{V}^\top$) when the two CNNs are set as the same. However, the outer product operation results in extremely high dimensional features, \ie, the bilinear pooled feature is reshaped into a vector $\bm{z} = {\rm vec}(\bm{V}\bm{V}^\top)\in \mathcal{R}^{d^2}$. This results in a large increase in the number of parameters in the classification module of the deep network, which can cause overfitting and make it impractical for realistic applications, especially for large-scale ones. 
To address this problem, Gao~\etal~\cite{compactBCNN} applied Tensor Sketch~\cite{tensorKDD13} to both approximate the second-order statistics of the original bilinear pooling operation and reduce feature dimensions.
Kong~\etal~\cite{lowrankBCNN} adopted a low-rank approximation to the covariance matrix and further learned a low-rank bilinear classifier. The resulting classifier can be evaluated without explicitly computing the bilinear feature matrix which results in a large reduction on the parameter size. Li~\etal~\cite{facBCNNiccv17} also modeled pairwise feature interaction by performing a quadratic transformation with a low-rank constraint. Yu~\etal~\cite{hbp18eccv} used a dimension reduction projection before bilinear pooling to alleviate dimension explosion. Zheng~\etal~\cite{deepbcnnNIPS19} applied bilinear pooling to feature channel groups where the bilinear transformation is represented by calculating pairwise interactions within each group. This also results in large saving in computation cost.

Beyond these approaches, some methods attempt to capture much higher-order (more than second-order) interactions of features to generate stronger and more discriminative feature representations. Cui~\etal~\cite{yinkernel17} introduced a kernel pooling method that captures arbitrarily ordered and non-linear features via compact feature mapping. 
Cai~\etal~\cite{highorleizhangiccv17} proposed a polynomial kernel based predictor to model higher-order statistics of convolutional activations across multiple layers for modeling part interactions. Subsequently, DeepKSPD~\cite{deepKSPDeccv18} was developed to jointly learn the deep local descriptors and the kernel-matrix based covariance representation in an end-to-end trainable manner.

As $\ell_2$ feature normalization can suppress the common patterns of high response and thereby enhance those discriminative features (\ie, the visual burstiness problem~\cite{grassmannpool,fvIJCV13}), the aforementioned  bilinear pooling based methods typically perform element-wise square root normalization followed by $\ell_2$-normalization on covariance matrix to improve performance. 
However, merely employing $\ell_2$-normalization can cause unstable high-order information and also lead to slow convergence. 
To this end, many methods have explored non-linearly scaling based on the  singular value decomposition (SVD) or eigendecomposition (EIG) to obtain more stability for second-order representations. Specifically, Li~\etal~\cite{isiccv17qilong} proposed to apply the power exponent to the eigenvalues of bilinear features to achieve better recognition accuracy. G$^2$DeNet~\cite{g2denetcvpr} further combined complementary first-order and second-order information via a Gaussian embedding and matrix square root normalization. 
iSQRT-COV~\cite{matsrnormcvpr18} and the improved B-CNN~\cite{improvebcnn17} used the Newton-Schulz iteration to approximate matrix square-root normalization with only matrix multiplication to decrease training time.
Recently, MOMN~\cite{MOMNtip2020} was proposed to simultaneously normalize a bilinear representation in terms of square-root, low-rank, and sparsity all within a multi-objective optimization framework.

\subsubsection{Designing Specific Loss Functions}

Loss functions play an important role in the construction of deep networks. 
They can directly affect both the learned classifiers and features. 
Thus, designing fine-grained tailored loss functions is an important direction for fine-grained image recognition.

Distinct from generic image recognition, in fine-grained classification, where samples that belong to different classes can be visually very similar, it is reasonable to prevent the classifier from being too confident in its outputs (\ie, discourage low entropy). Following this intuition, \cite{maximumEntro} also maximized the entropy of the output probability distribution when training networks on fine-grained tasks. Similarly, Dubey~\etal~\cite{pariECCV18} used a pairwise confusion optimization procedure to solve both overfitting and sample-specific artifacts in fine-grained recognition by bringing the different class-conditional probability distributions closer together and confusing the deep network. This allows a reduction in prediction over-confidence, therefore resulting in improved generalization performance.

Humans can effectively identify contrastive clues by comparing image pairs, and this type of metric / contrastive learning is also common in fine-grained recognition. 
Specifically, Sun~\etal~\cite{MAMCeccv} first learned multiple part-corresponding attention regions and then leveraged metric learning for pulling same-attention same-class features closer, while pushing different-attention or different-class features away. 
Furthermore, their approach can coherently enforce the correlations among different object parts during training. 
CIN~\cite{CINAAAI2020} pulls positive pairs closer while pushing negative pairs away via a contrastive channel interaction module which also exploits channel correlations between samples. 
API-Net~\cite{pairinterQiaoAAAI20} was also built upon a metric learning framework, and can adaptively discover contrastive cues from a pair of images and distinguish them via pairwise attention based interactions.

Designing a single loss function for localizing part-level patterns and further aggregating image-level representations has also been explored in the literature. 
Specifically, Sun~\etal~\cite{lingshaoaaai20ac} developed a gradient-boosting based loss function along with a diversification block to force the network to move swiftly to discriminate the hard classes. 
Concretely, the diversification block suppresses the discriminative regions of the class activation maps, and hence the network is forced to find alternative informative features. 
The gradient-booting loss focuses on difficult (\ie, confusing) classes for each image and boosts their gradient. 
MC-Loss~\cite{mclossTIP20} encourages the feature channels to be more discriminative by focusing on various part-level regions. 
They propose a single loss that does not require any specific network modifications for partial localization of fine-grained objects.

These aforementioned loss function based fine-grained recognition methods are backbone-agnostic and their performance can typically be improved by using more powerful backbone network architectures.

\subsubsection{Other Methods}

Beyond modelling the interactions between higher-order features and designing novel loss functions, another set of approaches involve constructing fine-grained tailored auxiliary tasks for obtaining unified and discriminative image representations.

BGL~\cite{BGLCVPR2016} was proposed to incorporate rich bipartite-graph labels into CNN training to model the important relationships among fine-grained classes. DCL~\cite{destructioncvpr19} performed a ``destruction and construction'' process to enhance the difficulty of recognition to guide the network to focus on discriminative parts for fine-grained recognition (\ie, by destruction learning) and then model the semantic correlation among parts of the object (\ie, by construction learning). Similar to DCL, Du~\etal~\cite{FGjigsaw20} tackled fine-grained representation learning using a jigsaw puzzle generator proxy task to encourage the network to learn at different levels of granularity and simultaneously fuse features at these levels together. Recently, a more direct fine-grained feature learning method~\cite{ganfgcvpr2020} was formulated with the goal of generating \emph{identity-preserved} fine-grained images in an adversarial learning manner to directly obtain a unified fine-grained image representation. 
The authors showed that this direct feature learning approach not only  preserved the identity of the generated images, but also significantly boosted the visual recognition performance in other challenging tasks like fine-grained few-shot learning~\cite{pcmFSFG}.

\subsection{Recognition with External Information}\label{sec:exter}

\begin{table*}[t!]
\scriptsize
\caption{Comparison of fine-grained  ``recognition with external information''  (cf. Section~\ref{sec:exter}) on multiple fine-grained benchmark datasets, including Birds (\emph{CUB200-2011}~\cite{WahCUB200_2011}), Dogs (\emph{Stanford Dogs}~\cite{Khosla11stanforddogs}), Cars (\emph{Stanford Cars}~\cite{cars}), and Aircrafts (\emph{FGVC Aircraft}~\cite{airplanes}). ``External info.'' denotes which kind of external information is used by the respective approach. ``Train anno.'' and ``Test anno.'' indicate the supervision used during training and testing, and ``--'' means the results are unavailable.}
\vspace{-1em}
\label{table:bigtable_recog2}
\setlength{\tabcolsep}{3.3pt}
\begin{tabular}{|l|l|c|c|c|c|c|c|c||c|c|c|c|}
\hline
\multicolumn{3}{|c|}{\multirow{2}{*}{Methods}}                                                                                                                                                                                            & \multirow{2}{*}{Published in} & \multirow{2}{*}{Train anno.} & \multirow{2}{*}{Test anno.} & \multicolumn{1}{c|}{\multirow{2}{*}{External info.}} & \multicolumn{1}{c|}{\multirow{2}{*}{Backbones}} & \multicolumn{1}{c||}{\multirow{2}{*}{Img. resolution}} & \multicolumn{4}{c|}{Accuracy}                                                                                            \\ \cline{10-13} 
\multicolumn{3}{|c|}{}                                                                                                                                                                                                                    & \multicolumn{1}{c|}{}                              & \multicolumn{1}{c|}{}                             & \multicolumn{1}{c|}{}                            & \multicolumn{1}{c|}{}                                & \multicolumn{1}{c|}{}                           & \multicolumn{1}{c||}{}                                 & \multicolumn{1}{c|}{\textit{Birds}}  & \multicolumn{1}{c|}{\textit{Dogs}}   & \multicolumn{1}{c|}{\textit{Cars}}   & \multicolumn{1}{c|}{\textit{Aircrafts}} \\ \hline\hline
\multicolumn{1}{|l|}{\multirow{16}{*}{\rotatebox{90}{\tiny Fine-grained recognition with external information}}} & \multicolumn{1}{l|}{\multirow{8}{*}{\rotatebox{90}{\tiny With web / auxiliary data}}} & \multicolumn{1}{c|}{HAR-CNN~\cite{augregCVPR2015}}      & \multicolumn{1}{c|}{CVPR 2015}                     & \multicolumn{1}{c|}{BBox}                         & \multicolumn{1}{c|}{BBox}                        & \multicolumn{1}{c|}{Web data}                        & \multicolumn{1}{c|}{Alex-Net}                   & \multicolumn{1}{c||}{$224\times 224$}                  & \multicolumn{1}{c|}{--}     & \multicolumn{1}{c|}{49.4\%} & \multicolumn{1}{c|}{80.8\%} & \multicolumn{1}{c|}{--}        \\ \cline{3-13} 
\multicolumn{1}{|l|}{}                                                               & \multicolumn{1}{l|}{}                                                                    & \multicolumn{1}{c|}{Xu~\etal~\cite{zhetaoICCV2015}}     & \multicolumn{1}{c|}{ICCV 2015}                     & \multicolumn{1}{c|}{BBox+Parts}                   & \multicolumn{1}{c|}{}                            & \multicolumn{1}{c|}{Web data}                        & \multicolumn{1}{c|}{CaffeNet}                   & \multicolumn{1}{c||}{$224\times 224$}                  & \multicolumn{1}{c|}{84.6\%} & \multicolumn{1}{c|}{--}     & \multicolumn{1}{c|}{--}     & \multicolumn{1}{c|}{--}        \\ \cline{3-13} 
\multicolumn{1}{|l|}{}                                                               & \multicolumn{1}{l|}{}                                                                    & \multicolumn{1}{c|}{Krause~\etal~\cite{krause2016unreasonable}}     & \multicolumn{1}{c|}{ECCV 2016}                     & \multicolumn{1}{c|}{}                   & \multicolumn{1}{c|}{}                            & \multicolumn{1}{c|}{Web data}                        & \multicolumn{1}{c|}{Inception-v3}                   & \multicolumn{1}{c||}{$224\times 224$}                  & \multicolumn{1}{c|}{92.3\%} & \multicolumn{1}{c|}{80.8\%}     & \multicolumn{1}{c|}{--}     & \multicolumn{1}{c|}{93.4\%}        \\ \cline{3-13} 
\multicolumn{1}{|l|}{}                                                               & \multicolumn{1}{l|}{}                                                                    & \multicolumn{1}{c|}{Niu~\etal~\cite{webZSL2018cvpr}}    & \multicolumn{1}{c|}{CVPR 2018}                     & \multicolumn{1}{c|}{}                             & \multicolumn{1}{c|}{}                            & \multicolumn{1}{c|}{Web data}                        & \multicolumn{1}{c|}{VGG-16}                     & \multicolumn{1}{c||}{$224\times 224$}                  & \multicolumn{1}{c|}{76.5\%} & \multicolumn{1}{c|}{85.2\%} & \multicolumn{1}{c|}{--}     & \multicolumn{1}{c|}{--}        \\ \cline{3-13} 
\multicolumn{1}{|l|}{}                                                               & \multicolumn{1}{l|}{}                                                                    & \multicolumn{1}{c|}{MetaFGNet~\cite{eccvmetalearning}}  & \multicolumn{1}{c|}{ECCV 2018}                     & \multicolumn{1}{c|}{}                             & \multicolumn{1}{c|}{}                            & \multicolumn{1}{c|}{Auxiliary data}                  & \multicolumn{1}{c|}{ResNet-34}                  & \multicolumn{1}{c||}{$224\times 224$}                  & \multicolumn{1}{c|}{87.6\%} & \multicolumn{1}{c|}{96.7\%} & \multicolumn{1}{c|}{--}     & \multicolumn{1}{c|}{--}        \\ \cline{3-13} 
\multicolumn{1}{|l|}{}                                                               & \multicolumn{1}{l|}{}                                                                    & \multicolumn{1}{c|}{Xu~\etal~\cite{weblyTPAMI18zhe}}    & \multicolumn{1}{c|}{IEEE TPAMI 2018}                    & \multicolumn{1}{c|}{BBox+Parts}                   & \multicolumn{1}{c|}{}                            & \multicolumn{1}{c|}{Web data}                        & \multicolumn{1}{c|}{Alex-Net}                   & \multicolumn{1}{c||}{$224\times 224$}                  & \multicolumn{1}{c|}{84.6\%} & \multicolumn{1}{c|}{--}     & \multicolumn{1}{c|}{--}     & \multicolumn{1}{c|}{--}        \\ \cline{3-13} 
\multicolumn{1}{|l|}{}                                                               & \multicolumn{1}{l|}{}                                                                    & \multicolumn{1}{c|}{Yang~\etal~\cite{webjufengTIP18}}   & \multicolumn{1}{c|}{IEEE TIP 2018}                      & \multicolumn{1}{c|}{}                             & \multicolumn{1}{c|}{}                            & \multicolumn{1}{c|}{Web data}                        & \multicolumn{1}{c|}{ResNet-50}                  & \multicolumn{1}{c||}{$224\times 224$}                  & \multicolumn{1}{c|}{--}     & \multicolumn{1}{c|}{87.4\%} & \multicolumn{1}{c|}{--}     & \multicolumn{1}{c|}{--}        \\ \cline{3-13} 
\multicolumn{1}{|l|}{}                                                               & \multicolumn{1}{l|}{}                                                                    & \multicolumn{1}{c|}{Sun~\etal~\cite{xiaxiaoaaai19}}     & \multicolumn{1}{c|}{AAAI 2019}                     & \multicolumn{1}{c|}{}                             & \multicolumn{1}{c|}{}                            & \multicolumn{1}{c|}{Web data}                        & \multicolumn{1}{c|}{ResNet-50}                  & \multicolumn{1}{c||}{$224\times 224$}                  & \multicolumn{1}{c|}{--}     & \multicolumn{1}{c|}{87.1\%} & \multicolumn{1}{c|}{--}     & \multicolumn{1}{c|}{--}        \\ \cline{3-13} 
\multicolumn{1}{|l|}{}                                                               & \multicolumn{1}{l|}{}                                                                    & \multicolumn{1}{c|}{Zhang~\etal~\cite{webyazhouAAAI20}} & \multicolumn{1}{c|}{AAAI 2020}                     & \multicolumn{1}{c|}{}                             & \multicolumn{1}{c|}{}                            & \multicolumn{1}{c|}{Web data}                        & \multicolumn{1}{c|}{VGG-16}                     & \multicolumn{1}{c||}{$224\times 224$}                  & \multicolumn{1}{c|}{77.2\%} & \multicolumn{1}{c|}{--}     & \multicolumn{1}{c|}{78.7\%} & \multicolumn{1}{c|}{72.9\%}    \\ \cline{2-13} 
\multicolumn{1}{|l|}{}                                                               & \multicolumn{1}{l|}{\multirow{7}{*}{\rotatebox{90}{\tiny With multi-modal data}}}           & \multicolumn{1}{c|}{CVL~\cite{yuxinpengcvpr2017}}       & \multicolumn{1}{c|}{CVPR 2017}                     & \multicolumn{1}{c|}{}                             & \multicolumn{1}{c|}{}                            & \multicolumn{1}{c|}{Language texts}                  & \multicolumn{1}{c|}{VGG-16}                     & \multicolumn{1}{c||}{Not given}                        & \multicolumn{1}{c|}{85.6\%} & \multicolumn{1}{c|}{--}     & \multicolumn{1}{c|}{--}     & \multicolumn{1}{c|}{--}        \\ \cline{3-13} 
\multicolumn{1}{|l|}{}                                                               & \multicolumn{1}{l|}{}                                                                    & \multicolumn{1}{c|}{Zhang~\etal~\cite{audioaaai2018}}   & \multicolumn{1}{c|}{AAAI 2018}                     & \multicolumn{1}{c|}{BBox}                         & \multicolumn{1}{c|}{}                            & \multicolumn{1}{c|}{Audio}                           & \multicolumn{1}{c|}{VGG-16}                     & \multicolumn{1}{c||}{$227\times 227$}                  & \multicolumn{1}{c|}{85.6\%} & \multicolumn{1}{c|}{--}     & \multicolumn{1}{c|}{--}     & \multicolumn{1}{c|}{--}        \\ \cline{3-13} 
\multicolumn{1}{|l|}{}                                                               & \multicolumn{1}{l|}{}                                                                    & \multicolumn{1}{c|}{Zhang~\etal~\cite{audioaaai2018}}   & \multicolumn{1}{c|}{AAAI 2018}                     & \multicolumn{1}{c|}{BBox}                         & \multicolumn{1}{c|}{BBox}                        & \multicolumn{1}{c|}{Audio}                           & \multicolumn{1}{c|}{VGG-16}                     & \multicolumn{1}{c||}{$227\times 227$}                  & \multicolumn{1}{c|}{86.6\%} & \multicolumn{1}{c|}{--}     & \multicolumn{1}{c|}{--}     & \multicolumn{1}{c|}{--}        \\ \cline{3-13} 
\multicolumn{1}{|l|}{}                                                               & \multicolumn{1}{l|}{}                                                                    & \multicolumn{1}{c|}{T-CNN~\cite{TCNNijcai18}}           & \multicolumn{1}{c|}{IJCAI 2018}                    & \multicolumn{1}{c|}{BBox}                         & \multicolumn{1}{c|}{}                            & \multicolumn{1}{c|}{Knowledge base + Texts}          & \multicolumn{1}{c|}{ResNet-50}                  & \multicolumn{1}{c||}{$224\times 224$}                  & \multicolumn{1}{c|}{86.5\%} & \multicolumn{1}{c|}{--}     & \multicolumn{1}{c|}{--}     & \multicolumn{1}{c|}{--}        \\ \cline{3-13} 
\multicolumn{1}{|l|}{}                                                               & \multicolumn{1}{l|}{}                                                                    & \multicolumn{1}{c|}{KERL~\cite{KERLijcai18}}            & \multicolumn{1}{c|}{IJCAI 2018}                    & \multicolumn{1}{c|}{}                             & \multicolumn{1}{c|}{}                            & \multicolumn{1}{c|}{Knowledge base}                  & \multicolumn{1}{c|}{VGG-16}                     & \multicolumn{1}{c||}{$224\times 224$}                  & \multicolumn{1}{c|}{87.0\%} & \multicolumn{1}{c|}{--}     & \multicolumn{1}{c|}{--}     & \multicolumn{1}{c|}{--}        \\ \cline{3-13} 
\multicolumn{1}{|l|}{}                                                               & \multicolumn{1}{l|}{}                                                                    & \multicolumn{1}{c|}{PMA~\cite{PMATIPkaisong}}           & \multicolumn{1}{c|}{IEEE TIP 2020}                      & \multicolumn{1}{c|}{}                             & \multicolumn{1}{c|}{}                            & \multicolumn{1}{c|}{Language texts}                  & \multicolumn{1}{c|}{VGG-16}                     & \multicolumn{1}{c||}{$448\times 448$}                  & \multicolumn{1}{c|}{88.2\%} & \multicolumn{1}{c|}{--}     & \multicolumn{1}{c|}{--}     & \multicolumn{1}{c|}{--}        \\ \cline{3-13} 
\multicolumn{1}{|l|}{}                                                               & \multicolumn{1}{l|}{}                                                                    & \multicolumn{1}{c|}{PMA~\cite{PMATIPkaisong}}           & \multicolumn{1}{c|}{IEEE TIP 2020}                      & \multicolumn{1}{c|}{}                             & \multicolumn{1}{c|}{}                            & \multicolumn{1}{c|}{Language texts}                  & \multicolumn{1}{c|}{ResNet-50}                  & \multicolumn{1}{c||}{$448\times 448$}                  & \multicolumn{1}{c|}{88.7\%} & \multicolumn{1}{c|}{--}     & \multicolumn{1}{c|}{--}     & \multicolumn{1}{c|}{--}        \\ \hline      
\end{tabular}
\end{table*}

Beyond the conventional recognition paradigms, which are restricted to using supervision associated with the images themselves, another paradigm is to leverage external information, \eg, web data, multi-modal data, or human-computer interactions, to further assist fine-grained recognition.

\subsubsection{Noisy Web Data}
Large and well-labeled training datasets are necessary in order to identify subtle differences between various fine-grained categories. However, acquiring accurate human labels for fine-grained categories is difficult due to the need for domain expertise and the myriads of fine-grained categories (\eg, potentially more than tens of thousands of subordinate categories in a meta-category). As a result, some fine-grained recognition methods seek to utilize freely available, but noisy, web data to boost recognition performance. The majority of existing work in this line can be roughly grouped into two directions.

The first direction involves scraping noisy labeled web data for the categories of interest as training data, which is regarded as \emph{webly supervised learning}~\cite{bohancvpr17,xiaxiaoaaai19,weblyTPAMI18zhe}. These approaches typically concentrate on: 1) overcoming the domain gap between easily acquired web images and the well-labeled data from standard datasets; and 2) reducing the negative effects caused by the noisy data. For instance, HAR-CNN~\cite{augregCVPR2015} utilized easily annotated meta-classes inherent in the fine-grained data and also acquired a large number of meta-class-labeled images from the web to regularize the models for improving recognition accuracy in a multi-task manner (\ie, for both the fine-grained and the meta-class data recognition task). Xu~\etal~\cite{zhetaoICCV2015} investigated if fine-grained web images could provide weakly-labeled information to augment deep features and thus contribute to robust object classifiers by building a multi-instance (MI) learner, \ie, treating the image as the MI bag and the proposal part bounding boxes as the instances of MI. Krause~\etal~\cite{krause2016unreasonable} introduced an alternative approach to combine a generic classification model with web data by excluding images that appear in search results for more than one category to combat cross-domain noise. Inspired by adversarial learning~\cite{GAN}, \cite{xiaxiaoaaai19} proposed an adversarial discriminative loss to encourage representation coherence between standard and web data.

The second direction is to transfer the knowledge from auxiliary categories with well-labeled training data to the test categories, which usually employs zero-shot learning~\cite{niulicvpr18} or meta learning~\cite{metalearnsurvey}. Niu~\etal~\cite{niulicvpr18} exploited zero-shot learning to transfer knowledge from annotated fine-grained categories to other fine-grained categories. 
Subsequently, Zhang~\etal~\cite{eccvmetalearning}, Yang~\etal~\cite{webjufengTIP18}, and Zhang~\etal~\cite{webyazhouAAAI20} investigated different approaches for selecting high-quality web training images to expand the training set. 
Zhang~\etal~\cite{eccvmetalearning} proposed a novel regularized meta-learning objective to guide the learning of network parameters so they are optimal for adapting to the target fine-grained categories. 
Yang~\etal~\cite{webjufengTIP18} designed an iterative method that progressively selects useful images by modifying the label assignment using multiple labels to lessen the impact of the labels from the noisy web data. 
Zhang~\etal~\cite{webyazhouAAAI20} leveraged the prediction scores in different training epochs to supervise the separation of useful and irrelevant noisy web images.

\subsubsection{Multi-Modal Data}\label{sec:multimodal}

Multi-modal analysis has attracted a lot of attention with the rapid growth of multi-media data, \eg, image, text, knowledge bases, etc. 
In fine-grained recognition, multi-modal data can be used to establish joint-representations/embeddings by incorporating multi-modal data in order to boost fine-grained recognition accuracy. 
Compared with strong semantic supervision from fine-grained images (\eg, part annotations), text descriptions are a weak form of supervision (\ie, they only provide image-level supervision). 
One advantage however, is that text descriptions can be relatively accurately generated by non-experts. 
Thus, they are both easy and cheap to be collected. In addition, high-level knowledge graphs, when available, can contain rich knowledge (\eg, \emph{DBpedia}~\cite{DBpedia15}). In practice, both text descriptions and knowledge bases are useful extra guidance for advancing fine-grained image representation learning.

\begin{figure}[t!]
\centering
{\includegraphics[width=0.95\columnwidth]{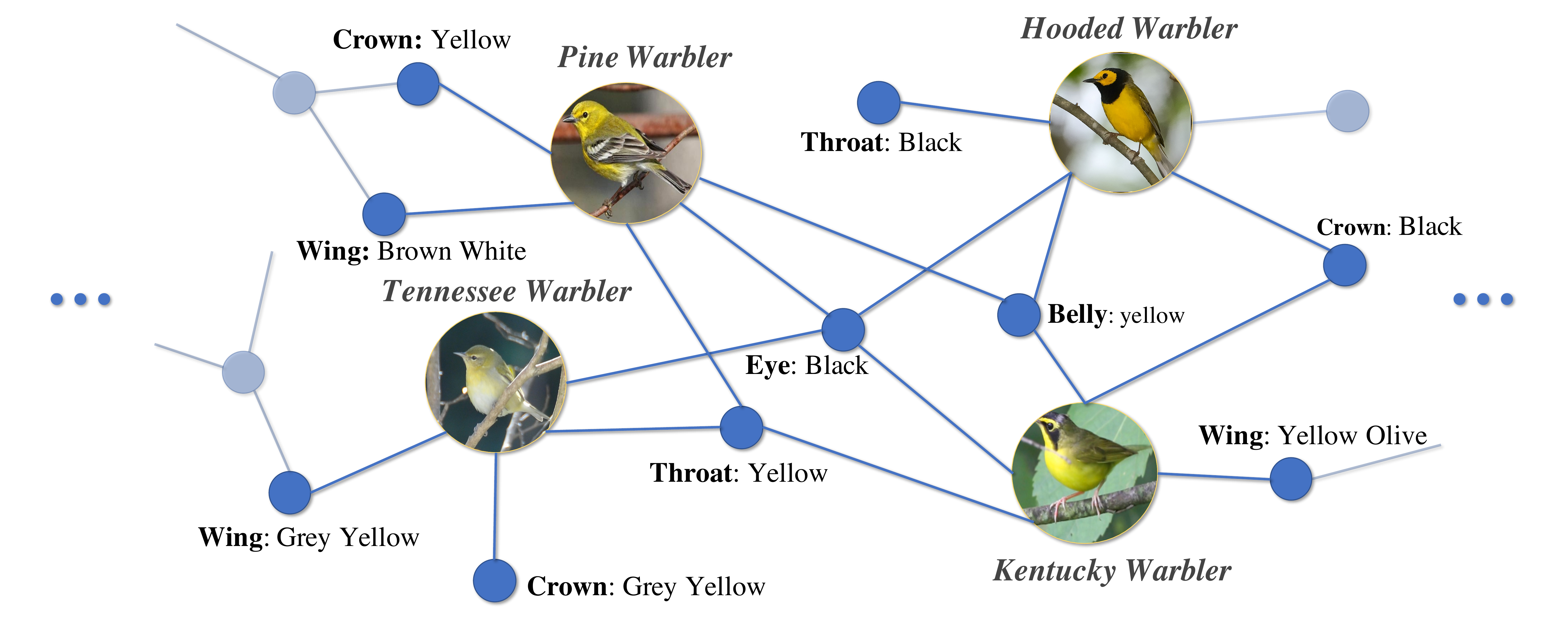}}
\vspace{-1.5em}
\caption{An example knowledge graph for modeling category-attribute correlations in \emph{CUB200-2011}~\cite{WahCUB200_2011}.}
\label{fig:KB}
\end{figure}

Reed~\etal~\cite{fgtextcvpr2016} collected text descriptions, and introduced a structured joint embedding for zero-shot fine-grained image recognition by combining text and images. 
Later, He and Peng~\cite{yuxinpengcvpr2017} combined vision and language bi-streams in a joint end-to-end fashion to preserve the intra-modality and inter-modality information for generating complementary fine-grained representations. 
Later, PMA~\cite{PMATIPkaisong} proposed a mask-based self-attention mechanism to capture the most discriminative parts in the visual modality. In addition, they explored using out-of-visual-domain knowledge using language with query-relational attention. 
Multiple PMA blocks for the vision and language modalities were aggregated and stacked using the proposed progressive mask strategy. 
For fine-grained recognition with knowledge bases, some existing works ~\cite{KERLijcai18,TCNNijcai18} have introduced  knowledge base information (using attribute label associations, cf. Figure~\ref{fig:KB}) to implicitly enrich the embedding space, while also reasoning about the discriminative attributes of fine-grained objects. 
Concretely, T-CNN~\cite{TCNNijcai18} explored using semantic embeddings from knowledge bases and text, and then trained a CNN to linearly map image features to the semantic embedding space to aggregate multi-modal information. 
To incorporate the knowledge representation into image features, KERL~\cite{KERLijcai18} employed a gated graph network to  propagate node messages through the graph to generate the knowledge representation. 
Finally, \cite{audioaaai2018} incorporated audio information related to the  fine-grained visual categories of interest to boost recognition accuracy.

\subsubsection{Humans-in-the-Loop}

Human-in-the-loop methods~\cite{hilJAIR} combine the complementary strengths of human knowledge with computer vision algorithms. 
Fine-grained recognition with humans-in-the-loop is typically posed in an iterative fashion and requires the vision system to be intelligent about when it queries the human for assistance. 
Generally, for these kinds of recognition methods,  in each round the system is seeking to understand how humans perform recognition, \eg, by asking expert humans to label the image class~\cite{yincvpr16}, or by identifying key part localization and selecting discriminative features~\cite{wisdomtpami16} for fine-grained recognition.

\subsection{Summary and Discussion}

The \emph{CUB200-2011}~\cite{WahCUB200_2011}, \emph{Stanford Dogs}~\cite{Khosla11stanforddogs}, \emph{Stanford Cars}~\cite{cars}, and \emph{FGVC Aircraft}~\cite{airplanes} benchmarks are among the most influential datasets in fine-grained recognition. 
Tables~\ref{table:bigtablerecog1} and \ref{table:bigtable_recog2} summarize results achieved by the fine-grained methods belonging to three recognition learning paradigms outlined above, \ie, ``recognition by localization-classification subnetworks'', ``recognition by end-to-end feature encoding'', and ``recognition with external information''. 
A chronological overview can be seen in Figure~\ref{fig:detailedParadigm}. 
The main observations can be summarized as follows:
\begin{itemize}[leftmargin = 2.2em]
\item There is an explicit correspondence between the reviewed methods and the aforementioned challenges of fine-grained recognition.
Specifically, the challenge of capturing subtle visual differences can be overcome by  localization-classification methods (cf. Section~\ref{sec:loc-cls}) or via specific construction-based tasks~\cite{destructioncvpr19,FGjigsaw20,ganfgcvpr2020}, as well as human-in-the-loop methods. The challenge of characterizing fine-grained tailored features is alleviated by performing high-order feature interactions or by leveraging multi-modality data. Finally, the challenging nature of FGIA can be somewhat addressed by designing specific loss functions~\cite{maximumEntro,pariECCV18,lingshaoaaai20ac} for achieving better accuracy.

\begin{table}[t!]
\small
\centering
\caption{Comparative fine-grained recognition results on \emph{CUB200-2011} using different input image resolutions. The results in this table are conducted based on a vanilla ResNet-50 trained at the respective resolution.}
\vspace{-1em}
\setlength{\tabcolsep}{4.2pt}
\label{table:resolution}
\begin{tabular}{|c|c|c|c|c|}
\hline
Resolution & $224\times 224$    & $280\times 280$     & $336\times 336$     & $392\times 392$     \\ \hline\hline
Accuracy   & 81.6\% & 83.3\% & 85.0\% & 85.6\% \\ \hline
\end{tabular}
\end{table}

\item Among the different learning paradigms, the ``recognition by localization-classification subnetworks'' and ``recognition by end-to-end feature encoding'' paradigms are the two most frequently investigated ones.
\item Part-level reasoning of fine-grained object categories boosts fine-grained recognition accuracy especially for non-rigid objects, \eg, birds. Modeling the internal semantic interactions/correlations among discriminative parts has attracted increased attention in recent years, cf.~\cite{hsnetcvpr17,yizhoucvpr19,graphAAAI20,FDLAAAI2020,s3niccv2019,MAMCeccv,gaussCVPR20fg,CINAAAI2020}.
\item Non-rigid fine-grained object recognition (\eg, birds or dogs) is more challenging than rigid fine-grained objects (\eg, cars or aircrafts), which is partly due to the variation on object appearance.
\item Fine-grained image recognition performance improves as image resolution increases \cite{cui2018large}. 
Comparison results on \emph{CUB200-2011} of different image resolutions are reported in Table~\ref{table:resolution}. 
\item There is a trade-off between recognition and localization ability for the ``recognition by localization-classification subnetworks'' paradigm, 
which might impact a single integrated network's recognition accuracy. 
Such a trade-off is also reflected in practice when trying to achieve better recognition results, in that training usually involves alternating optimization of the two networks or separately training the two followed by joint tuning. 
Alternating or multistage strategies complicate the tuning of the integrated network.%
\item While effective, most end-to-end encoding networks are less human-interpretable and less consistent in their accuracy across non-rigid and rigid visual domains compared to localization-classification subnetworks. Recently, it has been observed that several higher-order pooling methods attempt to understand such kind of methods by presenting visual interpretation~\cite{bcnnTPAMI} or from an optimization perspective~\cite{glocovQL20}.
\item ``Recognition by localization-classification subnetworks'' based methods are challenging to apply when the fine-grained parts are not consistent across the meta-categories (\eg, \emph{iNaturalist}~\cite{inat2017}). 
Here, unified end-to-end feature encoding methods are more appropriate.
\end{itemize}

\section{Fine-Grained Image Retrieval}\label{sec:fgretrieval}

Fine-grained retrieval is another fundamental aspect of FGIA that has gained more traction in recent years. 
What distinguishes fine-grained retrieval from fine-grained recognition is that in addition to estimating the sub-category correctly, it is also necessary to rank all the instances so that images belonging to the same sub-category are ranked highest based on the fine-grained details in the query. Specifically, in fine-grained retrieval we are given a database of images of the same meta-category (\eg, birds or cars) and a query, and the goal is to return images related to the query based on relevant fine-grained features. 
Compared to generic image retrieval, which focuses on retrieving near-duplicate images based on similarities in their content (\eg, texture, color, and shapes), fine-grained retrieval focuses on retrieving the images of the same category type (\eg, the same subordinate species of animal or the same model of vehicle). 
What makes it more challenging is that objects of fine-grained categories have exhibit subtle differences, and can vary in pose, scale, and orientation or can contain large cross-modal differences (\eg, in the case of sketch-based retrieval).

\begin{figure*}[t!]
\centering
{\includegraphics[width=0.9\textwidth]{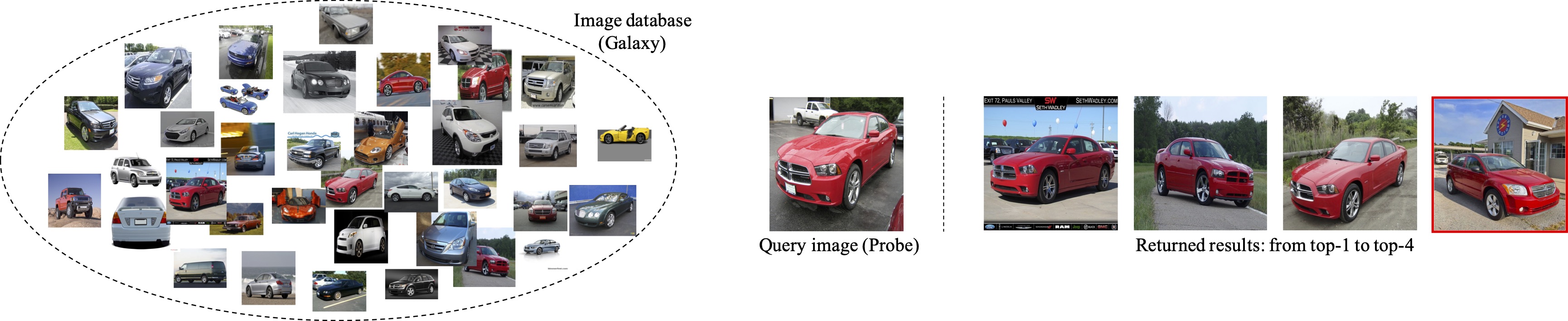}}
\vspace{-10pt}
\caption{An illustration of fine-grained content-based image retrieval (FG-CBIR). Given a query image (\emph{aka} probe) depicting a ``\texttt{Dodge Charger Sedan 2012}'', fine-grained retrieval is required to return images of the same car model from a car database (\emph{aka} galaxy). In this figure, the fourth returned image, marked with a red outline, is incorrect as it is a different car model, it is a ``\texttt{Dodge Caliber Wagon 2012}''.}
\label{fig:retrievaldemo}
\end{figure*}

\begin{table*}[t!]
\scriptsize
\centering
\caption{Comparison of recent fine-grained content-based image retrieval methods on \textit{CUB200-2011}~\cite{WahCUB200_2011} and \textit{Stanford Cars}~\cite{cars}. \emph{Recall}@$K$ is the average recall over all query images in the test set.}
\vspace{-1em}
\setlength{\tabcolsep}{4.5pt}
\label{table:CBIR}
\begin{tabular}{cccccclllllll}
\hline
\multicolumn{1}{|c|}{\multirow{3}{*}{Methods}}        & \multicolumn{1}{c|}{\multirow{3}{*}{Published in}} & \multicolumn{1}{c|}{\multirow{3}{*}{Supervised}} & \multicolumn{1}{c|}{\multirow{3}{*}{Backbones}} & \multicolumn{1}{c||}{\multirow{3}{*}{Img. Resolution}} & \multicolumn{8}{c|}{Recall@$K$}                                                                                                                                                                                               \\ \cline{6-13} 
\multicolumn{1}{|c|}{}                                & \multicolumn{1}{c|}{}                              & \multicolumn{1}{c|}{}                             & \multicolumn{1}{c|}{}                           & \multicolumn{1}{c||}{}                                 & \multicolumn{4}{c|}{\textit{Birds}}                                                                                    & \multicolumn{4}{c|}{\textit{Cars}}                                                                                     \\ \cline{6-13} 
\multicolumn{1}{|c|}{}                                & \multicolumn{1}{c|}{}                              & \multicolumn{1}{c|}{}                             & \multicolumn{1}{c|}{}                           & \multicolumn{1}{c||}{}                                 & \multicolumn{1}{c|}{1}    & \multicolumn{1}{c|}{2}    & \multicolumn{1}{c|}{4}    & \multicolumn{1}{c|}{8}    & \multicolumn{1}{c|}{1}    & \multicolumn{1}{c|}{2}    & \multicolumn{1}{c|}{4}    & \multicolumn{1}{c|}{8}    \\ \hline\hline
\multicolumn{1}{|c|}{SCDA~\cite{Wei16scda}}           & \multicolumn{1}{c|}{TIP 2017}                      & \multicolumn{1}{c|}{}                             & \multicolumn{1}{c|}{VGG-16}                     & \multicolumn{1}{c||}{$224\times 224$}                  & \multicolumn{1}{c|}{62.2\%} & \multicolumn{1}{c|}{74.2\%} & \multicolumn{1}{c|}{83.2\%} & \multicolumn{1}{c|}{90.1\%} & \multicolumn{1}{c|}{58.5\%} & \multicolumn{1}{c|}{69.8\%} & \multicolumn{1}{c|}{79.1\%} & \multicolumn{1}{c|}{86.2\%} \\ \hline
\multicolumn{1}{|c|}{CRL-WSL~\cite{xiawuijcai18}}     & \multicolumn{1}{c|}{IJCAI 2018}                    & \multicolumn{1}{c|}{$\checkmark$}                 & \multicolumn{1}{c|}{VGG-16}                     & \multicolumn{1}{c||}{$224\times 224$}                  & \multicolumn{1}{c|}{65.9\%} & \multicolumn{1}{c|}{76.5\%} & \multicolumn{1}{c|}{85.3\%} & \multicolumn{1}{c|}{90.3\%} & \multicolumn{1}{c|}{63.9\%} & \multicolumn{1}{c|}{73.7\%} & \multicolumn{1}{c|}{82.1\%} & \multicolumn{1}{c|}{89.2\%} \\ \hline
\multicolumn{1}{|c|}{DGCRL~\cite{xiawuaaai19}}        & \multicolumn{1}{c|}{AAAI 2019}                     & \multicolumn{1}{c|}{$\checkmark$}                 & \multicolumn{1}{c|}{ResNet-50}                  & \multicolumn{1}{c||}{Not given}                        & \multicolumn{1}{c|}{67.9\%} & \multicolumn{1}{c|}{79.1\%} & \multicolumn{1}{c|}{86.2\%} & \multicolumn{1}{c|}{91.8\%} & \multicolumn{1}{c|}{75.9\%} & \multicolumn{1}{c|}{83.9\%} & \multicolumn{1}{c|}{89.7\%} & \multicolumn{1}{c|}{94.0\%} \\ \hline
\multicolumn{1}{|c|}{Zeng~\etal~\cite{pce2020imavis}} & \multicolumn{1}{c|}{Image and Vis. Comp. 2020}     & \multicolumn{1}{c|}{$\checkmark$}                 & \multicolumn{1}{c|}{ResNet-50}                  & \multicolumn{1}{c||}{$224\times 224$}                  & \multicolumn{1}{c|}{70.1\%} & \multicolumn{1}{c|}{79.8\%} & \multicolumn{1}{c|}{86.9\%} & \multicolumn{1}{c|}{92.0\%} & \multicolumn{1}{c|}{86.7\%} & \multicolumn{1}{c|}{91.7\%} & \multicolumn{1}{c|}{95.2\%} & \multicolumn{1}{c|}{97.0\%} \\ \hline                        
\end{tabular}
\end{table*}

\begin{figure}[t!]
\centering
{\includegraphics[width=0.95\columnwidth]{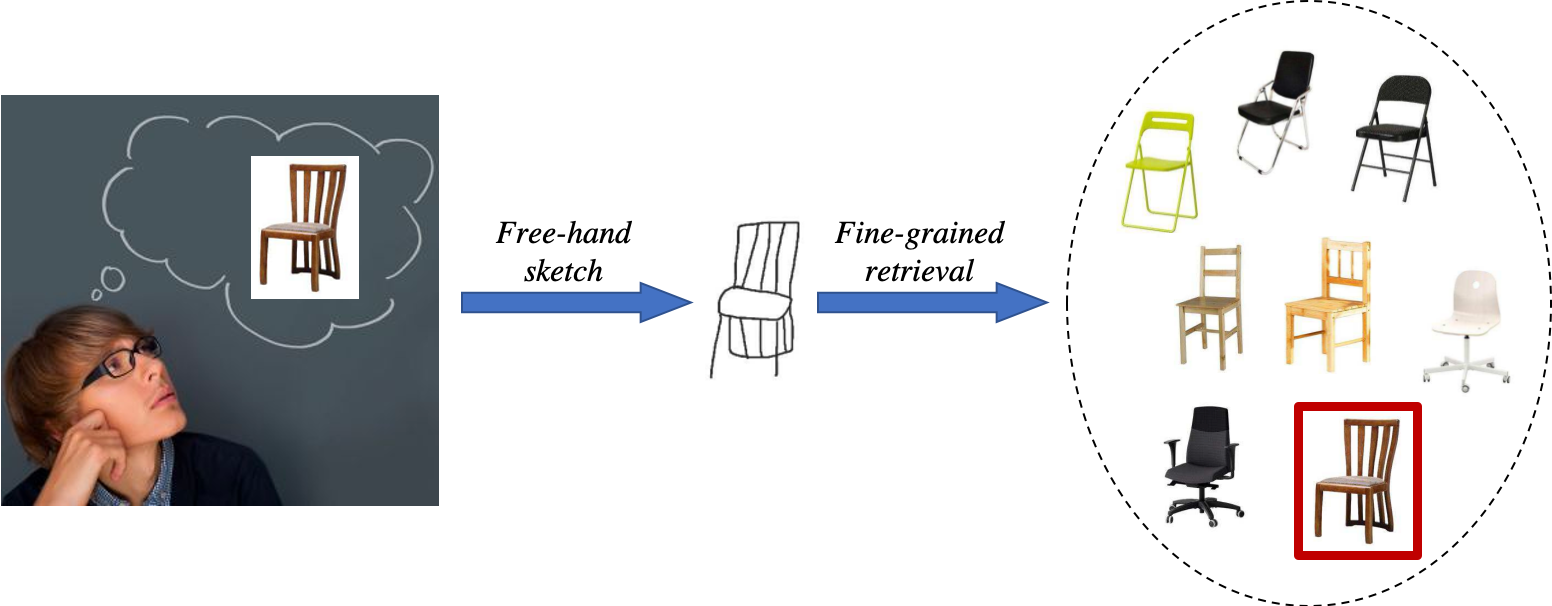}}
\vspace{-8pt}
\caption{An illustration of fine-grained sketch-based image retrieval (FG-SBIR), where a free-hand human sketch serves as the query for instance-level retrieval of images. FG-SBIR is challenging due to 1) the fine-grained and cross-domain nature of the task and 2) free-hand sketches are highly abstract, making fine-grained matching even more difficult.}
\label{fig:sketchdemo}
\end{figure}

Fine-grained retrieval techniques have been widely used in commercial applications, \eg, e-commerce (searching fine-grained products~\cite{retailvision}), touch-screen devices (searching fine-grained objects by sketches~\cite{SIGgraph16}), crime prevention (searching face photos~\cite{researchnote}), among others. Depending on the type of query image, the most studied areas of fine-grained image retrieval can be separated into two groups: fine-grained content-based image retrieval (FG-CBIR, cf. Figure~\ref{fig:retrievaldemo}) and fine-grained sketch-based image retrieval (FG-SBIR, cf. Figure~\ref{fig:sketchdemo}). 
Fine-grained image retrieval can also be expanded into fine-grained cross-media retrieval~\cite{pengacmmmcross}, which can utilize one media type to retrieve any media types, for example using an image to retrieve relevant text, video, or audio.

For performance evaluation, following the standard convention, FG-CBIR performance is typically measured using \emph{Recall@}$K$~\cite{recallatK} which is the average recall score over all $M$ query images in the test set. 
For each query, the top $K$ relevant images are returned. 
The recall score will be $1$ if there are at least one positive image in the top $K$ returned images, and $0$ otherwise. By formulation, the definition of \emph{Recall@}$K$ is as follows
\begin{equation}
\emph{Recall@}K = \frac{1}{M} \sum_{i=1}^M {\rm score_i}\,.
\end{equation}
For measuring FG-SBIR performance, \emph{Accuracy@}$K$ is commonly used, which is the percentage of sketches whose true-match photos are ranked in the top $K$:
\begin{equation}
\emph{Accuracy@}K = \frac{|I_{\rm correct}^K|}{K}\,,
\end{equation}
where $|I_{\rm correct}^K|$ is the number of true-match photos in top $K$.


\subsection{Content-based Fine-Grained Image Retrieval}

SCDA~\cite{Wei16scda} is one of the earliest examples of fine-grained image retrieval that used deep learning. 
It employs a pre-trained CNN to select meaningful deep descriptors by localizing the main object in an image without using explicit localization supervision. Unsurprisingly, they show that selecting only useful deep descriptors, by removing background features, can significantly benefit retrieval performance in such an unsupervised retrieval setting (\ie, requiring no image labels). Recently, supervised metric learning based approaches (\eg,~\cite{xiawuijcai18,xiawuaaai19}) have been proposed to overcome the retrieval accuracy limitations of unsupervised retrieval. These methods still include additional sub-modules specifically tailored for fine-grained objects, \eg, the weakly-supervised localization module proposed in~\cite{xiawuijcai18}, which is in turn inspired by ~\cite{Wei16scda}. 
CRL-WSL~\cite{xiawuijcai18} employed a centralized ranking loss with a weakly-supervised localization approach to train their feature extractor. DGCRL~\cite{xiawuaaai19} eliminated the gap between inner-product and the Euclidean distance in the training and test stages by adding a Normalize-Scale layer to enhance the intra-class separability and inter-class compactness with their Decorrelated Global-aware Centralized Ranking Loss. Recently, the Piecewise Cross Entropy loss~\cite{pce2020imavis} was proposed by modifying the traditional cross entropy function by reducing the confidence of the fine-grained model, which is similar to the basic idea of following the natural prediction confidence scores for fine-grained categories~\cite{maximumEntro,pariECCV18}.

The performance of recent fine-grained content-based image retrieval approaches are reported in Table~\ref{table:CBIR}. 
Although supervised metric learning based retrieval methods outperform their unsupervised counterparts, the absolute recall scores (\ie, \emph{Recall@}$K$) of the retrieval task still has room for further improvement. One promising direction is to integrate advanced techniques, \eg, attention mechanisms or higher-order feature interactions, which are successful for fine-grained recognition into FG-CBIR to achieve better retrieval accuracy. However, new large-scale FG-CBIR datasets are required to drive future progress, which are also desirable to be associated with other characteristics or challenges, \eg, open-world sub-category retrieval (cf. Section~\ref{sec:tax}).

\begin{table*}[t!]
\centering
\scriptsize
\caption{Comparison of fine-grained sketch-based image retrieval methods on  \textit{QMUL-Shoe}~\cite{thatshoecvpr}, \textit{QMUL-Chair}~\cite{thatshoecvpr}, \textit{QMUL-Handbag}~\cite{sketchretrievaliccv17}, \textit{Sketchy}~\cite{SIGgraph16}, and \textit{QMUL-Shoe-V2}~\cite{geneCVPRIR19}. \emph{Accuracy}@$K$ is the percentage of sketches whose true-match photos are ranked in the top $K$.}
\vspace{-1em}
\label{table:SBIR}
\begin{tabular}{|c|c||c|c|c|c|c|c|c|c|c|c|}
\hline
\multirow{3}{*}{Methods}          & \multirow{3}{*}{Published in} & \multicolumn{10}{c|}{Accuracy@$K$}                                                                                                                                                                         \\ \cline{3-12} 
                                  &                               & \multicolumn{2}{c|}{\emph{QMUL-Shoe}} & \multicolumn{2}{c|}{\emph{QMUL-Chair}} & \multicolumn{2}{c|}{\emph{QMUL-Handbag}} & \multicolumn{2}{c|}{\emph{Sketchy}} & \multicolumn{2}{c|}{\emph{QMUL-Shoe-V2}} \\ \cline{3-12} 
                                  &                               & 1                 & 10                & 1                  & 10                & 1                   & 10                 & 1                & 10               & 1                     & 10               \\ \hline\hline
Yu~\etal~\cite{thatshoecvpr}      & CVPR 2016                     & 39.1\%           & 87.8\%           & 69.1\%            & 97.9\%           & --                  & --                 & --               & --               & --                    & --               \\ \hline
Song~\etal~\cite{song2016deep}      & BMVC 2016                     & 50.5\%           & 91.3\%           & 78.3\%            & 98.9\%           & --                  & --                 & --               & --               & --                    & --               \\ \hline
Song~\etal~\cite{sketchretrievaliccv17} & ICCV 2017                     & 61.7\%           & 94.8\%           & 81.4\%            & 95.9\%           & 49.4\%             & 82.7\%            & --               & --               & --                    & --               \\ \hline
Li~\etal~\cite{TIPFGSBIR17}       & IEEE TIP 2017                 & 51.3\%           & 90.4\%           & 76.3\%            & 97.9\%           & --                  & --                 & 45.3\%          & 98.2\%          & --                    & --               \\ \hline
Radenovic~\etal~\cite{radenovic2018deep}      & ECCV 2018                    & 54.8\%           & 92.2\%           & 85.7\%            & 97.9\%           & 51.2\%                  & 85.7\%                 & --               & --               & --                    & --               \\ \hline
Zhang~\etal~\cite{zhang2018generative}      & ECCV 2018                    & 35.7\%           & 84.3\%           & 67.1\%            & 99.0\%           & --                  & --                 & --               & --               & --                    & --               \\ \hline
Pang~\etal~\cite{jigsawCVPR20}    & CVPR 2020                     & 56.5\%           & --                & 85.9\%            & --                & 62.9\%             & --                 & --               & --               & 36.5\%               & --               \\ \hline
Bhunia~\etal~\cite{bhunia2020sketch}    & CVPR 2020                     & --          & --          & --            & --          & --                  & --                 & --               & --               & --                    & 79.6\%                \\ \hline
Sain~\etal~\cite{sain2020cross}    & BMVC 2020                     &  --          &  --         &  --      &  --          & --                  & --                 & --               & --               & 36.3\%                     & 80.6\%               \\ \hline
\end{tabular}
\end{table*}

\subsection{Sketch-based Fine-Grained Image Retrieval}\label{sec:FGSBIR}

The goal of fine-grained sketch-based image retrieval (FG-SBIR) is to match specific photo instances using a free-hand sketch as the query modality. Distinct from the previously discussed content-based approach, the additional sketch-photo domain gap lies at the centre of FG-SBIR. Thus, the key is introducing a cross-modal representation that not only captures fine-grained characteristics present in the sketches, but also possesses the ability to traverse the sketch and image domain gap.

Existing FG-SBIR solutions generally aim to train a joint embedding space where sketches and photos can be compared in a nearest neighbor fashion. Specifically, Li~\etal~\cite{SBIRbmvc14} proposed the first method for FG-SBIR, where they learned a deformable part-based models (DPM) ~\cite{dpmCNN15} as a mid-level representation to discover and encode the various poses in the sketch and image domains independently. This was followed by a graph matching operation on the DPM to establish pose correspondences across the two domains. Yu~\etal~\cite{thatshoecvpr} first employed deep learning for this task, leveraging a triplet ranking model with a staged pre-training strategy to learn a joint embedding space for the two domains. This triplet ranking model was further augmented with auxiliary semantic attribute prediction and attribute ranking layers~\cite{song2016deep} to improve generalization~\cite{thatshoecvpr}. The implicit challenge encountered due to the large sketch-photo domain gap in FG-SBIR was tackled via a discriminative-generative hybrid model~\cite{cdfbirbmvc17} using cross-domain translation. 
Inspired on these approaches, Song~\etal~\cite{sketchretrievaliccv17} introduced an attention module and encoded the spatial position of visual features and combined both coarse and fine-grained semantic knowledge. 
Li~\etal~\cite{TIPFGSBIR17} introduced a part-aware learning approach to reduce the instance-level domain-gap by performing subspace alignment with  fine-grained attributes. While research on fine-grained SBIR has flourished over the years, a dilemma still remains -- whether sketching is a better input modality for fine-grained image retrieval compared to other alternatives, \eg, texts or attributes.
To answer this question, Song~\etal~\cite{song2017fine} showed the superiority of sketch over text for fine-grained retrieval, illustrating that each modality can complement the other when they are jointly modeled.

In contrast to these earlier works that were mostly based on Siamese-triplet networks, there have been several recent attempts to advance FG-SBIR performance. 
For instance, Pang~\etal~\cite{geneCVPRIR19} identified that a baseline triplet based model fails to generalize when exposed to unseen categories.
Thus, cross-category generalization was introduced through a domain generalization setup, where a universal manifold of prototypical visual sketches was modeled to dynamically represent the sketch/photo. On the other hand, Bhunia \etal~\cite{bhunia2020sketch} developed an on-the-fly retrieval setup via reinforcement learning that begins retrieving photos as soon as the user starts drawing.
ImageNet pre-trained weights have been considered as the de-facto choice for initializing sketch/photo embedding networks for FG-SBIR. However, following the recent progress in self-supervised learning~\cite{iclrUnsupRotation,moco,simclr}, Pang~\etal~\cite{jigsawCVPR20} performed a jigsaw solving strategy over mixed-modal patches between a particular photo and its edge-map. 
They showed this to be an effective pre-text task for self-supervised pre-training strategy and it improves FG-SBIR performance. 
Instead of performing independent sketch and photo embeddings, as in almost all previous works, Sain~\etal\cite{sain2020cross} used paired-embeddings by employing cross-modal co-attention and hierarchical stroke/region-wise feature fusion in order to deal with varying levels of sketch detail. 
In spite of a significant performance boost, it is noteworthy that paired-embeddings incur a significant computational overhead, by a multiple of at least the number of gallery photos, compared to other state-of-the-art methods. Recently, a new scene-level fine-grained SBIR task~\cite{scenesketcher} was explored. Liu~\etal~\cite{scenesketcher} proposed to utilize local features such as object instances and their visual detail, as well as global context (\eg, the scene layout), to deal with such a task.
 
In addition to the previous FG-SBIR methods, Zhang~\etal~\cite{zhang2018generative} and Radenovic~\etal~\cite{radenovic2018deep} also evaluated on popular FG-SBIR datasets, in addition to category-level SBIR ones. We compare the results of recent FG-SBIR methods in Table~\ref{table:SBIR}. Note that we only present methods whose evaluation strategies are uniform and consistent, and thus exclude those that involve problem specific evaluation protocols, such as zero-shot FG-SBIR~\cite{geneCVPRIR19}.





\section{Common Techniques Shared by both Fine-Grained Recognition and Retrieval}\label{sec:commontech}

The tasks of fine-grained image recognition and retrieval are complementary. As a result, there exists common techniques shared by both problems in the FGIA literature. In this section, we discuss these common techniques, in terms of methods and basic ideas, with the aim of motivating further work in these areas.

\textbf{Common Methods:} In the context of deep learning, both fine-grained recognition and retrieval tasks require discriminative deep feature embeddings to distinguish subtle differences between fine-grained objects. While recognition aims to distinguish category labels, retrieval aims to return an accurate ranking. To achieve these goals, deep metric learning and multi-modal matching can be viewed as two common sets of techniques that are applicable for both fine-grained recognition and retrieval.

Specifically, deep metric learning~\cite{metriccheck2020} attempts to map image data to an embedding space, where similar fine-grained images are close together and dissimilar images are far apart. In general, fine-grained recognition realizes metric learning by classification losses, where it includes a weight matrix to transform the embedding space into a vector of fine-grained class logits, \eg,~\cite{MAMCeccv,CINAAAI2020,pairinterQiaoAAAI20}. While, metric learning on fine-grained retrieval tasks (most cases without explicit image labels) is always achieved by means of embedding losses which operate on the relationships between fine-grained samples in a batch, \eg,~\cite{xiawuijcai18,xiawuaaai19}.

Recently, multi-modal matching methods~\cite{WACV2020fgictextvision,wacv2021visiontest} have emerged as powerful representation learning approaches to simultaneously boost fine-grained recognition and retrieval tasks. In particularly, Mafla~\etal~\cite{WACV2020fgictextvision} leveraged textual information along with visual cues to comprehend the existing intrinsic relation between the two modalities. \cite{wacv2021visiontest} employed a graph convolutional network to perform multi-modal reasoning and obtain relationship-enhanced multi-modal features. Such kind of methods reveal a new development trend of multi-modal learning in FGIA.

\textbf{Common Basic Ideas:} Beyond these common methods, many basic ideas are shared by both fine-grained recognition and retrieval, \eg, selecting useful deep descriptors~\cite{maskcnnPR,Wei16scda}, reducing the uncertainty of fine-grained predictions~\cite{maximumEntro,pariECCV18,pce2020imavis}, and deconstructing/constructing fine-grained images for learning fine-grained patterns~\cite{destructioncvpr19,jigsawCVPR20}. These observations further illustrate the benefit of consolidating work in fine-grained recognition and fine-grained retrieval in this survey paper.

\section{Future Directions}\label{sec:future}

Advances in deep learning have enabled significant progress in fine-grained image analysis (FGIA). 
Despite the success however, there are still many unsolved problems. Thus, in this section, we aim to explicitly point out these problems, and highlight some open questions to motivate future progression of the field. 

\textbf{Precise Definition of ``Fine-Grained'':} Although FGIA has existed for many years as a thriving sub-field of computer vision and pattern recognition, one fundamental problem in FGIA persists, \ie, the designation lacks a precise definition~\cite{cuiFGdefinition,song2020}. Specifically, the community always \emph{qualitatively} describes a so-called fine-grained dataset/problem as being ``fine-grained'' by roughly stating that its target objects belong to one meta-category. However, a precise definition of ``fine-grained'' could enable us to \emph{quantitatively} evaluate the granularity of a dataset. In addition, it would not only provide a better understanding of current algorithmic performance on tasks of different granularities, but also bring additional insights to the fine-grained community.

\textbf{Next-Generation Fine-Grained Datasets:} Classic fine-grained datasets such as \emph{CUB200-2011}, \emph{Stanford Dogs}, \emph{Stanford Cars}, and \emph{Oxford Flowers}, are overwhelmingly used for benchmarking performance in FGIA. 
However, by modern standards these datasets are relatively small-scale and are largely saturated in terms of performance. In the future, it would be valuable to see additional large-scale fine-grained datasets being promoted to replace these existing benchmarks, \eg, \emph{iNat2017}~\cite{inat2017}, \emph{Dogs-in-the-wild}~\cite{MAMCeccv}, \emph{RPC}~\cite{rpc}. State-of-the-art results on these datasets are 72.6\%~\cite{routingLTiclr21}, 78.5\%~\cite{MAMCeccv}, 80.5\%~\cite{dpnetacmmm}, respectively, which reveals the substantial room for further improvement.
Moreover, these new benchmarks should embrace and embody all the challenges associated with fine-grained learning in addition to being large-scale (in terms of both the number of sub-categories and images), contain diverse images captured in realistic settings, and include rich annotations. However, high-quality fine-grained datasets usually require domain-experts to annotate. 
This limits the development of fine-grained datasets to a certain extent. 
As a potential solution, constructing an unlabeled large-scale fine-grained database and employing unsupervised feature learning techniques (\eg, self-supervised learning~\cite{SSLsurvey}) could benefit discriminative features learning and further promote unsupervised downstream tasks~\cite{inat2021}. Also, synthetic data~\cite{syndatasurvey} is an increasingly popular tool for training deep models, and offers promising opportunities to be explored further for FGIA.


\textbf{Application to 3D Fine-Grained Tasks:} Most existing FGIA approaches target 2D fine-grained images and the value of 3D information (\eg, 3D pose labels or 3D shape information) is under-explored. 
How 3D object representations boost the performance of 2D FGIA approaches is an interesting and important problem. On the other hand, how 2D FGIA approaches generalize to 3D fine-grained applications~\cite{yushenTsinghua}, \eg, robotic bin picking, robot perception or augmented reality, is also worthy of future attention. To make progress in this area there are open questions associated with difficulties in obtaining accurate 3D annotations for many object categories, \eg, animals or other non-rigid objects.

\textbf{Robust Fine-Grained Representations:} One important factor which makes FGIA uniquely challenging is the overwhelming variability in real-world fine-grained images, including changes in viewpoint, object scale, pose, and factors such as part deformations, background clutter, and so on. Meanwhile, FGIA often dictates fine-grained patterns (derived from discriminative but subtle parts) to be identified to drive predictions, which makes it more sensitive to image resolutions, corruptions and perturbations, and adversarial examples. Despite advances in deep learning, current models are still exhibit a lack of robustness to these variations or disturbances, and this significantly constrains their usability in many real-world applications where high accuracy is essential. Therefore, how to effectively obtain robust fine-grained representations (\ie, not only containing discriminative fine-grained visual clues, but also resisting the interference of irrelevant information) requires further in-depth exploration. As discussed above, when coupled with next generation fine-grained datasets, there are questions related to the utility of self-supervised learning in these domains~\cite{cole2021does}. For example, will self-supervised learning help improving FGIA, and do we perform self-supervision on fine-grained data to generate robust fine-grained representations?


\textbf{Interpretable Fine-Grained Learning:} Unilaterally achieving higher accuracy compared to non-expert humans may no longer be the primary goal of FGIA. It is very important for fine-grained models to not only result in high accuracy but also to be interpretable~\cite{interpretabilitysurvey}. More interpretable FGIA could help the community to address several challenges when dealing with fine-grained tasks using deep learning, \eg, semantically debugging network representations, learning via human-computer communications at the semantic level, designing more effective, task-specific deep models, etc.

\textbf{Fine-Grained Few-Shot Learning:} Humans are capable of learning a new fine-grained concept with very little supervision (\eg, with a few image of a new species of bird), yet our best deep learning fine-grained systems need hundreds or thousands of labeled examples. Even worse, the supervision of fine-grained images are both time-consuming and expensive to obtain, as fine-grained objects need to be accurately labeled by domain experts. Thus, it is desirable to develop fine-grained few-shot learning (FGFS) methods~\cite{pcmFSFG,metaNIPSFSFG,cvprFGFS,su2020does}. The task of FGFS requires the learning system to build classifiers for novel fine-grained categories from few examples (\eg, less than 10). Robust FGFS methods could significantly strengthen the usability and scalability of fine-grained recognition.

\textbf{Fine-Grained Hashing:} Recently, larger-scale fine-grained datasets have been released, \eg,~\cite{Birdsnap14,deepfashion16,vegfru,inat2017,rpc,inat2021}. In real applications like fine-grained image retrieval, the computational cost of finding the exact nearest neighbor can be prohibitively high in cases where the reference database is very large. Image hashing~\cite{surveyhashtpami,wujunhashingijcai16} is a popular and effective technique for approximate nearest neighbor search, and has the potential to help with large-scale fine-grained data too. Therefore, targeting the big data challenge, fine-grained hashing~\cite{exchnet,FGhashTIP} is a promising direction worthy of further explorations.

\textbf{Automatic Fine-Grained Models:} Automated machine learning (AutoML)~\cite{automlnips} and neural architecture search (NAS)~\cite{nassurvey} have attracted growing attention of late. AutoML targets automating the process of applying machine learning to real-world tasks and  NAS is the process of automating neural network architecture design. Recent methods for AutoML and NAS could be comparable or even outperform hand-designed architectures in various computer vision applications. Thus, it is also logical that AutoML and NAS techniques could find better, and more tailor-made deep models for FGIA.

\textbf{Fine-Grained Analysis in More Realistic Settings:}\label{sec:realsetting} In the past decade, FGIA related techniques have been developed and have achieved good performance in standard computer vision benchmarks, \eg,~\cite{WahCUB200_2011,Khosla11stanforddogs,cars}. 
The vast majority of these existing FGIA benchmarks are however defined in a static and closed environment. One other big limitation of current FGIA datasets is that they typically contain large instances of the object (\ie, the objects of interest occupy most of the image frame). However, these settings are not representative of many real-world applications, \eg, recognizing retail products in storage racks by models trained with images collected in controlled environments~\cite{rpc} or recognizing/detecting tends of thousands of natural species in the wild~\cite{inat2017}. 
More research is needed in areas such as domain adaptation~\cite{fgdaICCV2017,FGDACVPR2020,WACV2020FGDA}, long-tailed distributions~\cite{cbfocal,bbn}, open world settings~\cite{openLTcvpr}, scale variations~\cite{inat2017}, fine-grained video understanding~\cite{zhu2018fine,munro2020multi}, knowledge transfer, and resource constrained embedded deployment, to name a few.

\section{Conclusion}\label{sec:conc}


We have presented a comprehensive survey on recent advances in deep learning based fine-grained image analysis (FGIA). 
Specifically, we advocated a broadened definition of FGIA, by consolidating work in fine-grained recognition and fine-grained retrieval. 
We enumerated gaps in existing research, pointed out a series of emerging topics, highlighted important future research directions, and illustrated that the problem of FGIA is still far from solved. 
However, given the significant improvements in performance over the past decade, we remain optimistic about future progress as we move towards more realistic and impactful applications.

\section*{Acknowledgments}

The authors would like to thank the editor and the anonymous reviewers for their constructive comments. This work was supported by Natural Science Foundation of Jiangsu Province of China under Grant (BK20210340), National Natural Science Foundation of China under Grant (61925201, 62132001, 61772256), the Fundamental Research Funds for the Central Universities (No. 30920041111), CAAI-Huawei MindSpore Open Fund, and ``111'' Program B13022.

\bibliographystyle{IEEEtran}
\bibliography{FGIA_survey_short}

\begin{IEEEbiography}[{\includegraphics[width=1in,height=1.25in,clip,keepaspectratio]{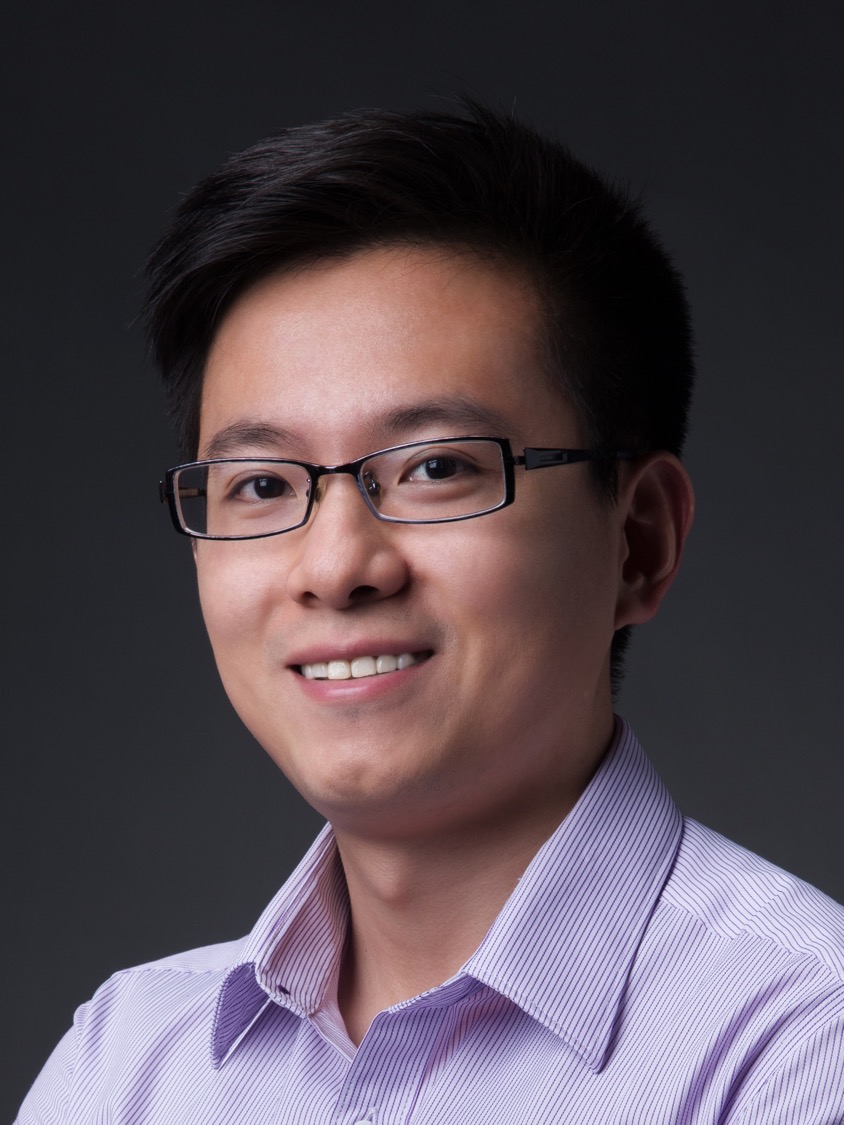}}]{Xiu-Shen Wei} is a Professor with the School of Computer Science and Engineering, Nanjing University of Science and Technology, China. He was a Program Chair for the workshops associated with ICCV, IJCAI, ACM Multimedia, etc. He has also served as a Guest Editor of Pattern Recognition Journal, and a Tutorial Chair for Asian Conference on Computer Vision (ACCV) 2022.
\end{IEEEbiography}

\begin{IEEEbiography}[{\includegraphics[width=1in,height=1.25in,clip,keepaspectratio]{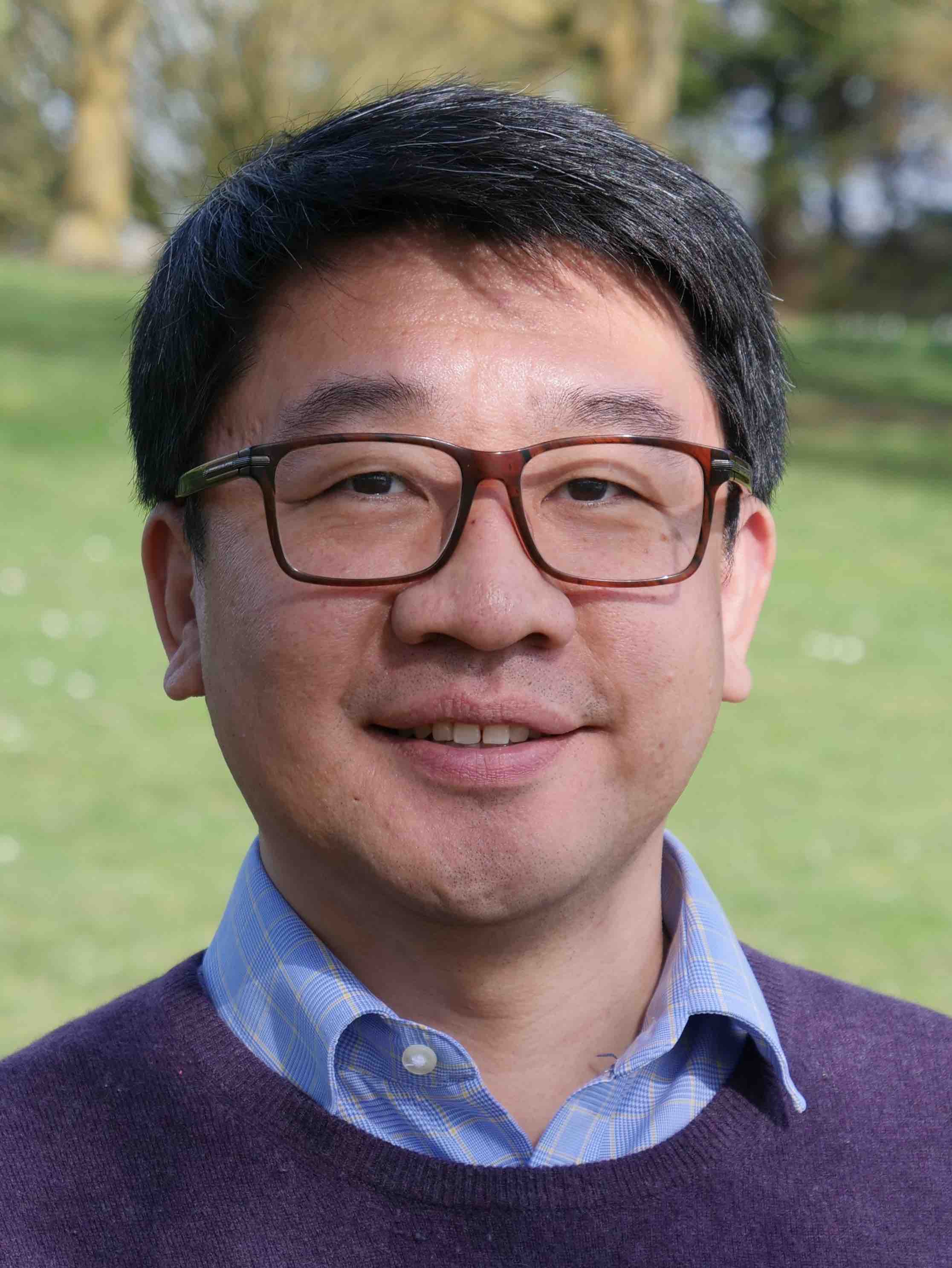}}]{Yi-Zhe Song} is a Chair Professor of Computer Vision and Machine Learning, and Director of SketchX Lab at the Centre for Vision Speech and Signal Processing (CVSSP), University of Surrey, UK. He is an Associate Editor of the IEEE Transactions on Pattern Analysis and Machine Intelligence (TPAMI), and a Program Chair for British Machine Vision Conference (BMVC) 2021.
\end{IEEEbiography}

\begin{IEEEbiography}[{\includegraphics[width=1in,height=1.25in,clip,keepaspectratio]{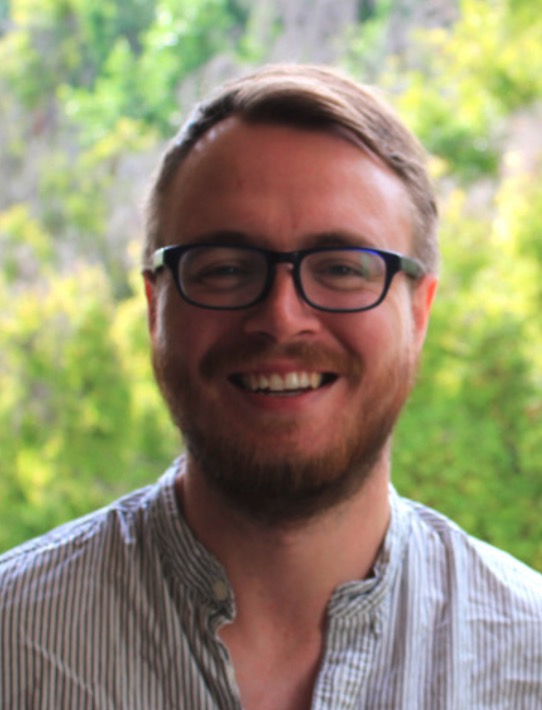}}]{Oisin Mac Aodha} is a Lecturer in Machine Learning at the University of Edinburgh, UK. 
He is a Fellow of the Alan Turing Institute and a European Laboratory for Learning and Intelligent Systems (ELLIS) Scholar. 
He was a Program Chair for the British Machine Vision Conference (BMVC) 2020.
\end{IEEEbiography}

\begin{IEEEbiography}[{\includegraphics[width=1in,height=1.25in,clip,keepaspectratio]{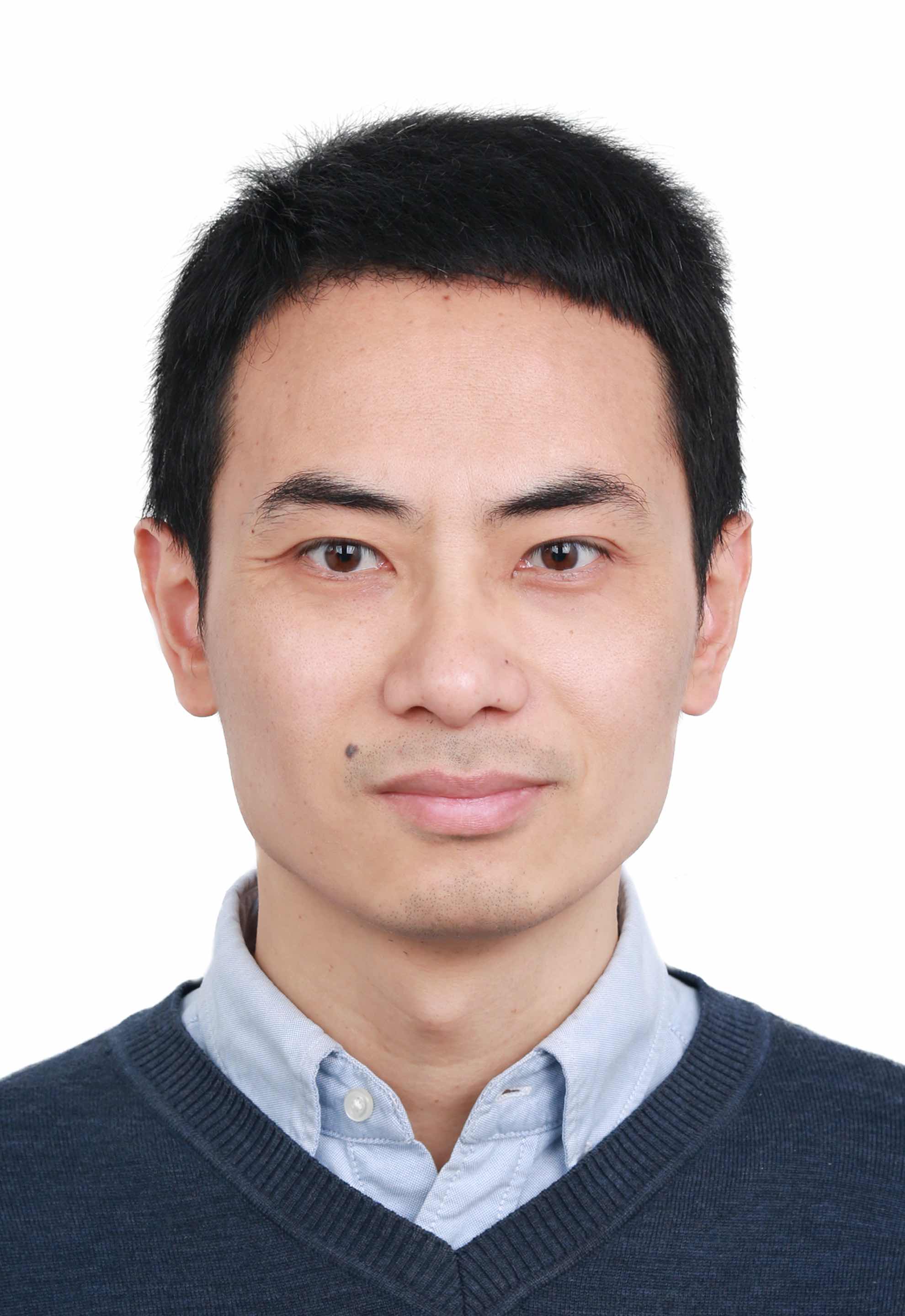}}]{Jianxin Wu} is currently a professor in the Department of Computer Science and Technology at Nanjing University, China, and is associated with the State Key Laboratory for Novel Software Technology, China.
\end{IEEEbiography}

\begin{IEEEbiography}[{\includegraphics[width=1in,height=1.25in,clip,keepaspectratio]{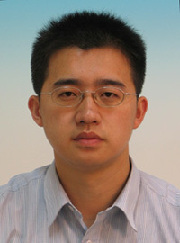}}]{Yuxin Peng} is currently the Boya Distinguished Professor with Wangxuan Institute of Computer Technology, Peking University, China.
\end{IEEEbiography}

\begin{IEEEbiography}[{\includegraphics[width=1in,height=1.25in,clip,keepaspectratio]{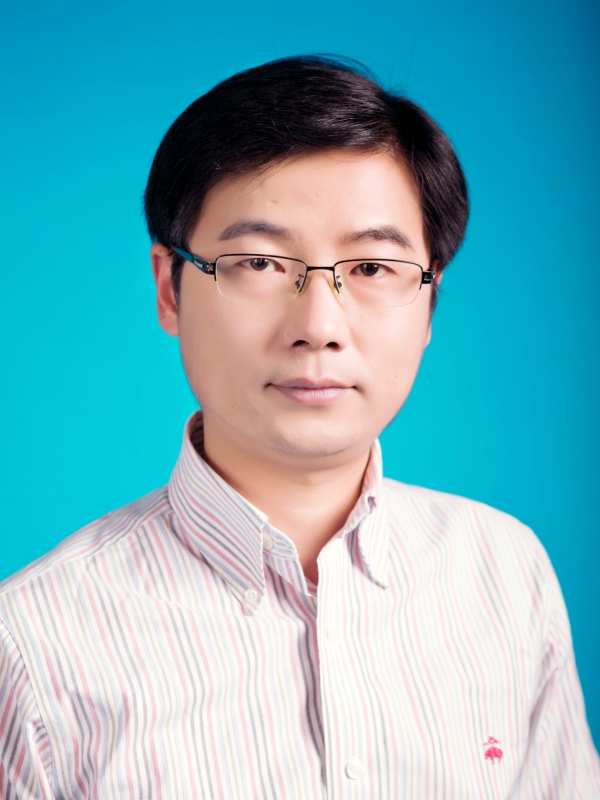}}]{Jinhui Tang} is a Professor with the School of Computer Science and Engineering, Nanjing University of Science and Technology, China.
\end{IEEEbiography}

\begin{IEEEbiography}[{\includegraphics[width=1in,height=1.25in,clip,keepaspectratio]{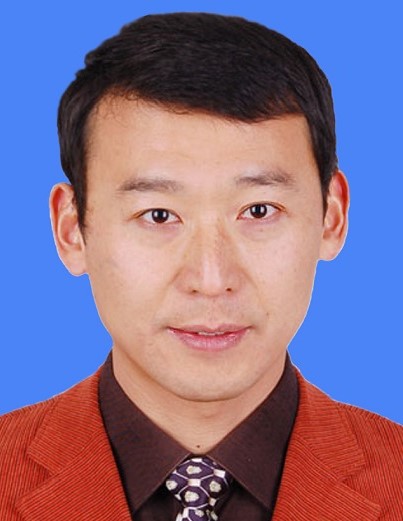}}]{Jian Yang} is a Chang-Jiang professor in the School of Computer Science and Technology of Nanjing University of Science and Technology, China.
\end{IEEEbiography}

\begin{IEEEbiography}[{\includegraphics[width=1in,height=1.25in,clip,keepaspectratio]{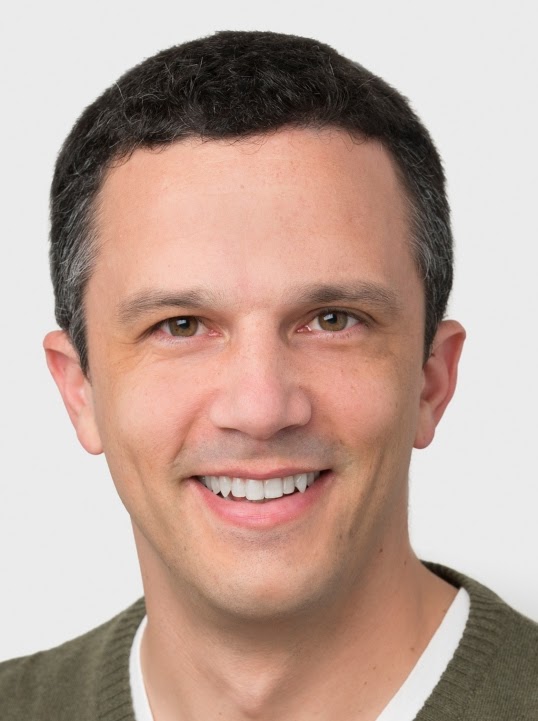}}]{Serge Belongie} is a professor of Computer Science at the University of Copenhagen (DIKU) and the director of the Pioneer Centre for AI, Denmark.
\end{IEEEbiography}

\end{document}